\documentclass{article} 
\usepackage{iclr2026_conference,times}

\usepackage{amsmath,amsfonts,bm}

\def\eqref#1{equation~\ref{#1}}

\def\1{\bm{1}}

\DeclareMathAlphabet{\mathsfit}{\encodingdefault}{\sfdefault}{m}{sl}
\SetMathAlphabet{\mathsfit}{bold}{\encodingdefault}{\sfdefault}{bx}{n}

\usepackage{url}
\usepackage{booktabs}       
\usepackage{amsfonts}       
\usepackage{nicefrac}       
\usepackage{microtype}      
\usepackage{xcolor}         
\usepackage{subcaption}
\newcommand{\methodName}{{\fontfamily{lmtt}\selectfont TBAR}}
\usepackage{graphicx} 
\usepackage{arydshln} 
\usepackage{multirow}
\usepackage{wrapfig}
\usepackage{amsmath}
\usepackage{adjustbox}
\usepackage{colortbl}
\usepackage{dsfont}
\usepackage{amsthm}
\newtheorem*{hypothesis}{Hypothesis}
\definecolor{mylightblue}{rgb}{0.122,0.467,0.706}
\usepackage[%
  colorlinks = true,
  citecolor  = mylightblue,
  linkcolor  = mylightblue,
  filecolor  = mylightblue,
  urlcolor   = mylightblue,
  unicode,     
  ]{hyperref}

\title{Backdoor Unlearning by Linear Task \\ Decomposition}

\author{
Amel Abdelraheem\thanks{Equal contribution.} \ \thanks{Corresponding author: \texttt{amel.abdelraheem@epfl.ch}} \ \textsuperscript{1}\quad
Alessandro Favero\footnotemark[1] \ \textsuperscript{1}\quad
Gérôme Bovet\textsuperscript{2}\quad
Pascal Frossard\textsuperscript{1}\\
\textsuperscript{1}EPFL, Lausanne, Switzerland \quad
\textsuperscript{2}Cyber-Defence Campus, armasuisse, Switzerland
}

\iclrfinalcopy 
\begin{document}

\maketitle

\begin{abstract}
Foundation models have revolutionized computer vision by enabling broad generalization across diverse tasks. Yet, they remain highly susceptible to adversarial perturbations and targeted backdoor attacks. Mitigating such vulnerabilities remains an open challenge, especially given that the large-scale nature of the models prohibits retraining to ensure safety. Existing backdoor removal approaches rely on costly fine-tuning to override the harmful behavior, and can often degrade performance on other unrelated tasks. This raises the question of whether backdoors can be removed without compromising the general capabilities of the models. In this work, we address this question and study how backdoors are encoded in the model weight space, finding that they are \textit{disentangled} from other benign tasks. Specifically, this separation enables the isolation and erasure of the backdoor's influence on the model with minimal impact on clean performance. Building on this insight, we introduce a simple unlearning method that leverages such disentanglement. Through extensive experiments with CLIP-based models and common adversarial triggers, we show that, given the knowledge of the attack, our method achieves approximately perfect unlearning, while retaining, on average, 96\% of clean accuracy. Additionally, we demonstrate that even when the attack and its presence are unknown, our method successfully unlearns backdoors by proper estimation using reverse-engineered triggers. Overall, our method consistently yields better unlearning and clean accuracy tradeoffs when compared to present state-of-the-art defenses.\looseness=-1
\end{abstract}


\section{Introduction} \label{sec:intro}

\looseness=-1 Foundation models have become a cornerstone of modern deep learning, offering broad generalization across a wide range of tasks through large-scale pre-training \citep{radford2021learning, jia2021scaling}. Among them, vision-language models like CLIP \citep{radford2021learning} play a fundamental role.
They not only demonstrate remarkable robustness to distribution shifts and \textit{zero-shot} performance on out-of-distribution benchmarks \citep{wortsman2022robust}, but their vision encoders also serve as a key component in many multimodal large language models, such as, e.g., LLaVA \citep{liu2023visual}.

\looseness=-1 However, the very success and widespread integration of these models make them a prime target for security threats, most notably \textit{backdoor attacks} \citep{carlini2021poisoning, bansal2023cleanclip} -- a class of threats that compromise model integrity even after training is complete. In a backdoor attack \citep{gu2017badnets}, an adversary poisons a small portion of the training data by embedding a fixed trigger pattern into inputs and mislabeling them to a target class. The resulting model appears to perform well on clean inputs but systematically misclassifies any input containing the trigger -- effectively granting the adversary precise control over model predictions. Such vulnerabilities pose a serious risk in safety-critical applications, including autonomous driving and medical diagnostics \citep{du2024defending,hanif2024baple}. 

\looseness=-1 Current defenses for CLIP largely fall into two categories: \textit{(i)} retraining the model from scratch using modified loss functions designed to resist backdoors, or \textit{(ii)} fine-tuning on clean data to override the malicious behavior \citep{bansal2023cleanclip,yang2024robust,goel2022cyclip}. However, full retraining is prohibitively expensive at scale, while fine-tuning -- though cheaper -- frequently induces \textit{catastrophic forgetting} \citep{french1999catastrophic}, whereby the pre-trained knowledge is erased. Furthermore, recent studies show that fine-tuning strategies struggle against more sophisticated attacks \citep{liang2024badclip}.

An alternative line of work, \textit{machine unlearning} \citep{cao2015towards}, seeks to selectively remove (or \textit{forget}) specific learned behaviors post-hoc, avoiding full retraining. Currently, the application of unlearning methods to targeted backdoor removal remains limited. Prominent unlearning algorithms such as \textit{gradient ascent} and its variants have been shown to fall short when applied to backdoor removal in small-scale settings \citep{pawelczyk2024machine}. Yet, their effectiveness in large-scale foundation models remains an open question.

\looseness=-1 In this paper, we introduce an efficient, post-hoc method for unlearning backdoors from vision-language foundation models while preserving their clean capabilities. Our approach builds on recent advances in model editing in weight space \citep{frankle2020linear,izmailov2018averaging,wortsman2021learning,wortsman2022model,rame2022diverse,ainsworth2022git,ilharco2022patching,ilharco2022editing}. 
In particular, \citet{ilharco2022editing} introduced the concept of a \textit{task vector}, which is the element-wise difference between the weights of a fine-tuned model and its pre-trained initialization. Task vectors provide a means to encode learned tasks as directions in weight space. They can be added to a model to inject functionality, subtracted to unlearn specific tasks, or combined to compose multi-task models. These manipulations are enabled by the \textit{disentanglement} of tasks in the weight space of pre-trained models, as recently formalized by \citet{ortiz2024task}. 

\begin{figure}[tb]
    \centering
    \includegraphics[width=0.9\linewidth]{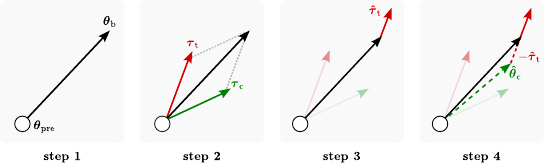}
    \caption{\looseness=-1 Backdoored models embed malicious behavior along with clean task performance. Instead of erasing all learned information, we propose a targeted approach: (1) Given a backdoored model, (2) the backdoor encodes two distinct directions, (3) fine-tuning the model on similarly constructed triggered data isolates the parameter shift associated with triggered information. (4) Negating this vector from the original parameters effectively removes the trigger while preserving clean task performance.}
    \label{fig:main}
\end{figure}

\looseness=-1 Motivated by these insights, we investigate how backdoors are encoded in the weight space of CLIP-based models. We find that weights can be linearly decomposed into clean and triggered components, effectively disentangling the malicious behavior from the model's benign capabilities. This disentanglement allows us to isolate the backdoor's influence by exploiting task arithmetic. In practice, this is achieved by fine-tuning the model on a small set of triggered examples to compute a ``trigger vector''. This vector isolates the malicious behavior and can thus be subtracted -- via task negation -- to surgically remove the backdoor while preserving clean model performance, as illustrated in Figure~\ref{fig:main}. We hence reframe the problem of backdoor unlearning as a simple problem of vector arithmetic.

Our main contributions are:
\begin{itemize}
    \item We leverage the weight disentanglement formalism to demonstrate that backdoors in CLIP-based transformer models are disentangled from clean knowledge in weight space, enabling targeted removal via linear operations without encountering catastrophic forgetting of non-adversarial knowledge.
    \item We introduce \methodName{} (\textbf{T}rigger removal by \textbf{B}ackdoor \textbf{AR}ithmetic), a lightweight approach for backdoor unlearning via weight-space task negation. When the trigger is known, \methodName{} unlearns ~99\% of the backdoor while retaining 96\% of obtained clean accuracy on average across \textit{(i)} image classification backdoor benchmarks and \textit{(ii)} large-scale image-captioning tasks. Notably, in the latter case, it outperforms state-of-the-art clean-data fine-tuning defenses while using less than 2\% of the data requirements.
    \item \looseness=-1 We extend \methodName{} to operate in large-scale settings in an attack-agnostic scenario by pairing it with reverse-engineered proxy triggers. Our method successfully sanitizes infected models, outperforming state-of-the-art defenses while preserving over 90\% clean accuracy.
\end{itemize}


\section{Problem Setup}\label{preliminary}
This work focuses on security vulnerabilities associated with backdoor attacks. Specifically, we consider the following threat model and defender assumptions. This setup is an extension of several previous settings in the literature \citep{carlini2021poisoning,bansal2023cleanclip,feng2023detecting,pawelczyk2024machine}.
\paragraph{Threat model} 
The adversary has full white-box access to a pre-trained model and fine-tuning data to be used to backdoor the model. The attack is conducted by injecting a small poisoned subset into a larger training dataset. The resulting backdoored model is released publicly and intended for downstream use by unaware users. Unless otherwise specified, we consider the attack successful if the triggered examples are predicted as a targeted label.

\paragraph{Defender assumptions}
\looseness=-1 The defender's goal is to remove the backdoor (i.e., reduce attack success rate to zero) while preserving the model's performance on clean data. The defender has full access to model weights. We consider two distinct practical scenarios: 
\begin{itemize}
    \item \textbf{Trigger-known:} 
    The defender is given a small forget set containing the true trigger, reflecting a common assumption in the context of backdoor defenses within unlearning studies, where an attack has been identified and its characteristics are known.
    \item \textbf{Trigger-unknown:} The defender does not know the true trigger but has access to a small set of clean data. 
\end{itemize} 

\section{Background}\label{WD_pre}

This section introduces the necessary tools for understanding model editing using weight interpolation. In particular, we recall the operation of task arithmetic and the property of weight disentanglement.

\paragraph{Notation}
Let a neural network be a parameterized function $f:\mathcal{X}\times\Theta\to\mathcal{Y}$ with inputs $x\in\mathcal{X}$ and weights $\boldsymbol{\theta}\in\Theta$. We identify a task $k\in[K]$ as a triplet $(\mathcal{D}_k,\mu_k,f_k^\star)$ with
domain $\mathcal{D}_k\subseteq\mathcal{X}$, input distribution $\mu_k$ ($\operatorname{supp}(\mu_k)=\mathcal{D}_k$), and
$f_k^\star:\mathcal{D}_k\to\mathcal{Y}$.

\paragraph{Model editing with task arithmetic} \citep{ilharco2022editing}
Finetuning a pre-trained model $\boldsymbol{\theta}_{\rm pre}$ on task $k$ yields new weights $\boldsymbol{\theta}_k^\star$. The change in weights $\boldsymbol{\tau}_k = \boldsymbol{\theta}_k^\star-\boldsymbol{\theta}_{\rm pre}$, defines the \textit{task vector}. Task arithmetic modifies the model by applying scaled task vectors: $\boldsymbol{\theta}_{\mathrm{new}} = \boldsymbol{\theta}_{\mathrm{pre}} + \alpha\,\boldsymbol{\tau}_k$ for a single task, or $\boldsymbol{\theta}_{\mathrm{new}} = \boldsymbol{\theta}_{\mathrm{pre}} + \sum_{k=1}^{K} \alpha_k\,\boldsymbol{\tau}_k$ for multiple tasks. The scalar coefficient $\alpha$ controls the strength of the edit as well as its direction, where positive values denote the learning of a task and negative values lead to the unlearning of the particular task.

\paragraph{Weight disentanglement} \citet{ortiz2024task} introduced \textit{weight disentanglement} as the property where the functional changes induced by a set of task vectors are localized to their respective task domains. Specifically, when multiple task vectors are linearly combined in weight space, the resulting model behaves as if it selectively applies each individual task's function only for inputs within that task's domain, reverting to the pre-trained model's behavior otherwise. The ability to perform task arithmetic with a set of task vectors $\mathcal{T}$ is a direct consequence of this weight disentanglement, where each task vector $\boldsymbol{\tau}_k$ encodes a distinct functional component specific to its domain $\mathcal{D}_k$. Formally, for a set of task vectors $\{\tau_k\}_{k \in [K]}$, the edited model satisfies weight disentanglement if (cf. \citet{ortiz2024task} for the formal definition):
\begin{align}\label{eq_WD}
f\!\left(x;\boldsymbol{\theta}_{\rm pre}+\sum_{k=1}^{K}\alpha_t\boldsymbol{\tau}_k\right)
= \sum_{k=1}^{K} f\!\left(x;\boldsymbol{\theta}_{\rm pre}+\alpha_k\boldsymbol{\tau}_k\right)\,\mathds{1}(x\in \mathcal{D}_k)
+ f\!\left(x;\boldsymbol{\theta}_{\rm pre}\right)\,\mathds{1}\!\left(x\notin \bigcup_{k=1}^{K} \mathcal{D}_k\right).
\end{align}
To measure the presence of weight disentanglement, \citet{ortiz2024task} introduced the weight disentanglement error, which measures the prediction disagreement between models obtained by applying the individual task vectors and the combination thereof, evaluated on the respective task supports. For two tasks, this reads:
\begin{equation}\label{wd_error_eq}
    \xi(\alpha_1, \alpha_2) = \sum_{i \in \{1, 2\}} \mathbb{E}_{x \sim \mu_i} \left[ \rm{dist} \left( f(x; \boldsymbol{\theta}_{\rm pre} + \alpha_i \boldsymbol{\tau}_i),\ f(x; \boldsymbol{\theta}_{\rm pre} + \alpha_1 \boldsymbol{\tau}_1 + \alpha_2 \boldsymbol{\tau}_2) \right) \right],
\end{equation}
where $\rm{dist}$ can be any distance metric between model outputs. For instance, for classification tasks ${\rm dist}(y_1, y_2) = \mathds{1}(y_1 \neq y_2)$.

In the next section, we study backdoor attacks through the lens of task arithmetic and weight disentanglement. We treat the benign task and the malicious backdoor behavior as two separate -- and, ideally, separable -- tasks operating on distinct data domains, i.e., clean or triggered inputs.

\section{\methodName: Trigger Removal by Backdoor Arithmetic}\label{sec:method}

\paragraph{Disentanglement of clean and triggered tasks}

Consider a model with pre-training weights $\boldsymbol{\theta}_{\text{pre}}$ that has been backdoored, resulting in weights $\boldsymbol{\theta}_{b}$. We investigate whether the joint training implicitly defines two tasks in parameter space, enabling the model’s behavior to decompose into clean and triggered components.  Formally, let $\boldsymbol{\tau}_c$ and $\boldsymbol{\tau}_t$ be the task vectors for the clean and triggered tasks, with domains $\mathcal{D}_c$ (clean images) and $\mathcal{D}_t$ (triggered images). Following the definition in Equation~\ref{eq_WD}, the backdoored model satisfies weight disentanglement with respect to these vectors if, $\forall x \in D_c \cup D_t$,
\begin{align}\label{WD_property}
    f\!\left(x;\boldsymbol{\theta}_{\rm pre}+\alpha_c\boldsymbol{\tau}_c+\alpha_t\boldsymbol{\tau}_t\right)
    = f\!\left(x;\boldsymbol{\theta}_{\rm pre}+\alpha_c\boldsymbol{\tau}_c\right)\,\mathds{1}(x\in D_c)
    + f\!\left(x,\boldsymbol{\theta}_{\rm pre}+\alpha_t\boldsymbol{\tau}_t\right)\,\mathds{1}(x\in D_t).
\end{align}

In this work, we formulate the following hypothesis:
\begin{hypothesis}
    The weights of vision foundation models satisfy weight disentanglement for common backdoor attacks, i.e., their output function $f$ satisfies Equation~\ref{WD_property}.
\end{hypothesis}

The crucial implication of this property is the existence of a specific direction in weight space, $\boldsymbol{\tau}_t$, that \textit{exclusively} governs the backdoor's malicious behavior. If this holds, removing the backdoor without causing catastrophic forgetting is possible: one simply needs to estimate $\boldsymbol{\tau}_t$ and subtract it from the model's weights. As we will demonstrate in the next section, this hypothesis holds in practice and allows us to effectively unlearn the backdoor without compromising the model's clean knowledge.

Provided this hypothesis holds, we only need to estimate the trigger vector in order to remove the attack. To accomplish this, we define a small, disjoint \textit{forget set} composed entirely of triggered image-target pairs. We fine-tune the suspected backdoored model $\boldsymbol{\theta}_b$ on this set, yielding updated weights $\boldsymbol{\theta}_{b+t}$. The parameter difference from this step gives us an estimate of the trigger direction:
\begin{equation}
    \hat{\boldsymbol{\tau}}_{t} = \boldsymbol{\theta}_{b+t} - \boldsymbol{\theta}_{b}
    \label{eq:trigger_vector_estimation}
\end{equation}
We can then surgically remove the backdoor's influence from the original backdoored model via task negation, yielding a cleaned model $\hat{\boldsymbol{\theta}}_{c}$:
\begin{equation}
    \hat{\boldsymbol{\theta}}_{c} = \boldsymbol{\theta}_{b} - \alpha \hat{\boldsymbol{\tau}}_{t}
    \label{eq:unlearning_op}
\end{equation}
where $\alpha$ is a scalar coefficient controlling the strength of the unlearning. We refer to this method as \textbf{T}rigger removal by \textbf{B}ackdoor \textbf{AR}ithmetic, or \methodName{}. Similarly with other weight interpolation techniques, we can use a small validation set for selecting the optimal value of the scaling coefficient $\alpha$ \citep{ilharco2022patching,ilharco2022editing,yadav2023ties,ortiz2024task,hazimehtask}.


\section{Trigger Vector Estimation with \methodName{}} \label{sec:analysis}

\looseness=-1 In this section, we focus on known-trigger settings, empirically validate our hypothesis, and show the effectiveness of \methodName{} on standard attacks. Moreover, we demonstrate that the learned \methodName{} vectors can be transferred across datasets and scale to practically relevant settings. 

\subsection{Disentanglement of clean and triggered Knowledge}

We start by following the standard model editing setup, where the CLIP text encoder stays frozen and only the visual encoder is fine-tuned \citep{wortsman2022robust,ilharco2022editing,yadav2023ties,ortiz2024task}. 
To construct a targeted poisoning attack on the visual encoder of CLIP by injecting triggered images into the training set, we follow \citep{carlini2021poisoning}. In particular, triggers are generated using three widely adopted methods: BadNet \citep{gu2017badnets}, which inserts a random square patch at a random location; Blended \citep{chen2017targeted}, which overlays uniform noise across the image; and WaNet \citep{nguyen2021wanet,qi2023towards}, which applies a subtle warping transformation. While BadNet represents a visible trigger, Blended and WaNet are considered invisible triggers. We evaluated three benchmark vision datasets: SUN397, CIFAR100, and ImageNet-1K, poisoned at a rate of 3\% of their training data. We report the per-dataset details in Appendix \ref{appendix_details}. To obtain the \methodName{} vectors, we use a small held-out forget set of 2000 examples from the training set and fine-tune using the same hyperparameter settings per dataset. Optimal scaling coefficients are found using a grid search, consistent with previous literature \citep{ilharco2022patching,ilharco2022editing,yadav2023ties,ortiz2024task,hazimehtask}.

\begin{table}[t]
\centering
\caption{Controlled experiments showing effectiveness of \methodName{} on single-task CLIP ViT-B/32 classifiers under three backdoor attacks. Clean Accuracy (CA $\uparrow$) and Attack Success Rate (ASR $\downarrow$) are reported before and after unlearning. Gray percentages denote CA retention and ASR removal relative to the backdoored model. Results are averaged over 4 seeds.}
\vspace{6pt}
{\fontsize{8pt}{9pt}\selectfont
\begin{tabular}{l l c|cc|cc}
\toprule
\multicolumn{3}{c|}{\cellcolor[HTML]{F7F7F7}\textbf{}} &
\multicolumn{2}{c|}{\cellcolor[HTML]{F7F7F7}\textbf{Attacked}} &
\multicolumn{2}{c}{\cellcolor[HTML]{F7F7F7}\textbf{\methodName{}}} \\
\textbf{Dataset} & \textbf{Attack} & \textbf{init\_CA} &
\textbf{CA $\uparrow$} & \textbf{ASR $\downarrow$} &
\textbf{CA $\uparrow$} & \textbf{ASR $\downarrow$} \\
\midrule
\multirow{3}{*}{\textit{SUN397}}
& BadNet  & 61.46 & 74.43 ± 0.34 & 91.40 ± 0.57 & 70.68 ± 0.84 {\color{gray}(94.96\%)} & 1.25 ± 2.37 {\color{gray}(98.63\%)} \\
& Blended & 61.46 & 74.72 ± 0.34 & 99.92 ± 0.12 & 73.36 ± 1.17 {\color{gray}(98.17\%)} & 0.00 ± 0.00 {\color{gray}(100\%)} \\
& WaNet   & 61.46 & 74.71 ± 0.12 & 99.62 ± 0.26 & 73.31 ± 0.40 {\color{gray}(98.13\%)} & 0.00 ± 0.00 {\color{gray}(100\%)} \\
\rowcolor[HTML]{F7F7F7} \multicolumn{7}{c}{} \\
\multirow{3}{*}{\textit{CIFAR100}}
& BadNet  & 62.46 & 88.77 ± 0.18 & 99.96 ± 0.04 & 85.61 ± 2.07 {\color{gray}(96.44\%)} & 0.02 ± 0.02 {\color{gray}(99.98\%)} \\
& Blended & 62.46 & 88.71 ± 0.22 & 99.98 ± 0.03 & 85.17 ± 1.96 {\color{gray}(96.01\%)} & 0.18 ± 0.48 {\color{gray}(99.82\%)} \\
& WaNet   & 62.46 & 88.66 ± 0.38 & 99.72 ± 0.05 & 87.61 ± 0.64 {\color{gray}(98.82\%)} & 0.04 ± 0.02 {\color{gray}(99.96\%)} \\
\rowcolor[HTML]{F7F7F7} \multicolumn{7}{c}{} \\
\multirow{3}{*}{\textit{ImageNet-1K}}
& BadNet  & 59.58 & 67.23 ± 0.18 & 93.56 ± 0.31 & 63.85 ± 0.29 {\color{gray}(94.97\%)} & 1.96 ± 2.38 {\color{gray}(97.91\%)} \\
& Blended & 59.58 & 67.50 ± 0.20 & 99.91 ± 0.04 & 66.06 ± 0.93 {\color{gray}(97.87\%)} & 0.00 ± 0.00 {\color{gray}(100\%)} \\
& WaNet   & 59.58 & 67.64 ± 0.18 & 99.86 ± 0.03 & 65.77 ± 1.20 {\color{gray}(97.24\%)} & 0.00 ± 0.00 {\color{gray}(100\%)} \\
\bottomrule
\end{tabular}
}
\label{tab:single-task-negation}
\end{table}

Table \ref{tab:single-task-negation} presents the full unlearning results across all datasets and attack types, reporting clean accuracy (CA) and attack success rate (ASR) before and after applying \methodName{}. \methodName{} consistently removes backdoors effectively, reducing ASR by over 98\% in all cases. Notably, this comes with only a moderate drop in obtained clean accuracy (i.e., 4\% on average), indicating that \methodName{} successfully isolates and removes triggered behavior from the model’s weights. 

\paragraph{Empirical validation of the disentanglement hypothesis}
We now validate our hypothesis and show that our successful unlearning is due to the disentanglement between clean and triggered behaviors by using the \textit{weight disentanglement error} $\xi$ (defined in Equation~\ref{wd_error_eq}). To do this, we must first construct the clean and trigger task vectors to be compared. Starting with $\hat{\boldsymbol{\tau}}_{\rm t}$ (from Equation~\ref{eq:trigger_vector_estimation}) as the estimated \textit{direction} of the trigger, we find an optimal scaling coefficient, $\alpha^{*}$, defined as the value that reduces the attack success rate to zero. This allows us to define the optimal trigger vector as $\alpha^{*}\hat{\boldsymbol{\tau}}_{\rm t}$. To define the corresponding clean vector, we first define the total update vector from pre-training to the backdoored state as $\boldsymbol{\tau}_{b} = \boldsymbol{\theta}_{b} - \boldsymbol{\theta}_{\rm pre}$. The clean vector, $\hat{\boldsymbol{\tau}}_{\rm c}$, is then computed as the residual of the total update: $\hat{\boldsymbol{\tau}}_{\rm c} = \boldsymbol{\tau}_{b} - \alpha^{*}\hat{\boldsymbol{\tau}}_{\rm t}$. If our disentanglement hypothesis holds, we expect to find a low disentanglement error between the resulting merged model and single models constructed using $\hat{\boldsymbol{\tau}}_{\rm c}$ and $\alpha^{*}\hat{\boldsymbol{\tau}}_{\rm t}$, on the respective data supports. 

\begin{figure}[tb]
    \centering
    \includegraphics[width=0.9\linewidth]{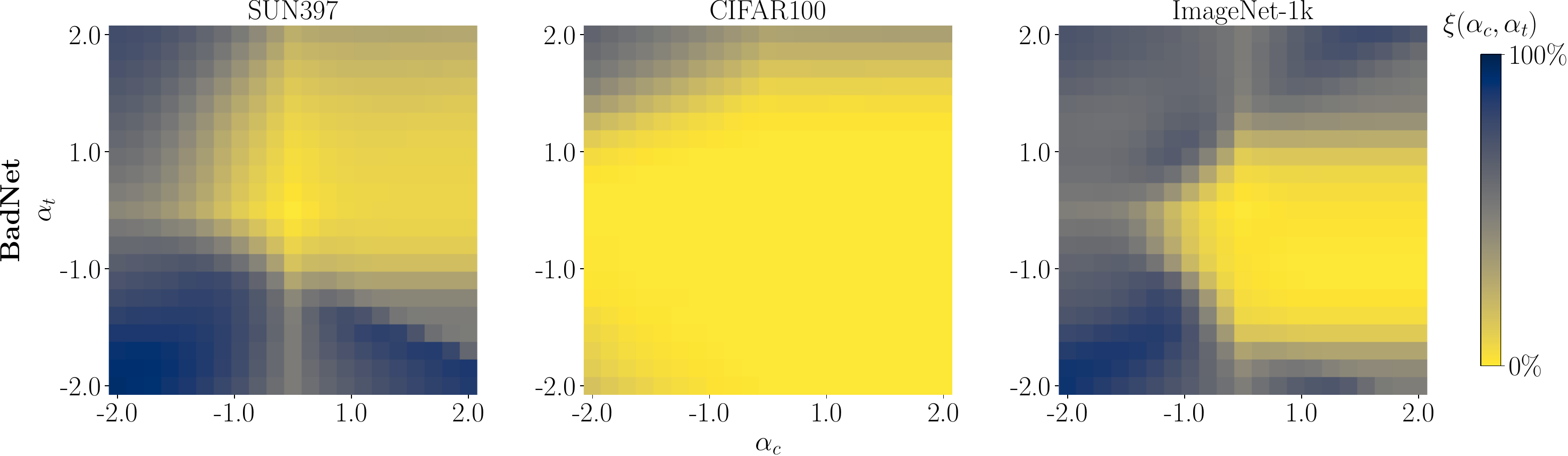}
    \caption{Weight disentanglement between clean and triggered tasks. We estimate the triggered direction $\boldsymbol{\hat \tau}_t$ from the backdoored model and define the clean direction $\boldsymbol{\hat\tau}_c$ as the residual after negation. The plots show the disentanglement error $\xi(\alpha_c, \alpha_t)$ between these task vectors, following \citet{ortiz2024task}. Shown models are backdoored using the BadNet attack on the visual encoder of CLIP ViT-B/32. Similar plots for the other attacks are provided in Appendix \ref{appendix_analysis}.}
    \label{fig:badnet disentanglment}
    \vspace{-1.5em}
\end{figure}

Visualizations of the weight disentanglement error presented in Figure~\ref{fig:badnet disentanglment} confirm the disentanglement in weight space and our hypothesis. In fact, the large bright regions at the center of the plots, indicating low error, show that the two tasks exhibit strong separation in weight space, providing evidence that triggered and clean vectors correspond to distinct directions\footnote{Notice that this is an analytical step and, in practice, it is not needed for our method's operation.}.

\subsection{Generalization and transferability of trigger vectors}

One of the main motivations behind using task vectors is their modularity: the ability to apply or combine them across models without retraining. In the case of backdoor unlearning, we therefore investigate a similar question: does a \methodName{} vector trained on one dataset capture the backdoor mechanism in a way that transfers to other models infected with the same attack?

\looseness=-1 Indeed, if the vector encodes only the trigger-to-misdirection behavior, rather than task-specific semantics, it should remain effective across models trained on different datasets, as long as the backdoor type and trigger remain consistent. To test it, we evaluate unlearning performance in out-of-distribution settings using vectors extracted from a backdoored ImageNet-1K model. We apply these vectors to remove backdoors from models trained with CIFAR100 and SUN397, respectively. 
\begin{wraptable}{r}{0.53\textwidth}
\centering
\caption{
Unlearning performance on CIFAR100 and SUN397 using \methodName{} vectors extracted using a backdoored ImageNet-1K model. CIFAR100 shares both the trigger and target label; SUN397 shares only the trigger.
}
\label{tab:transfer_w_in_full}
\begin{adjustbox}{width=\linewidth}
{\fontsize{8pt}{9pt}\selectfont
\begin{tabular}{@{}lllcc|cc@{}}
\toprule
& \textbf{} &
& \textbf{CA $\uparrow$} & \textbf{ASR $\downarrow$} 
& \textbf{CA (\methodName{}) $\uparrow$} & \textbf{ASR (\methodName{}) $\downarrow$} \\
\midrule

\rowcolor[HTML]{F7F7F7}
\multicolumn{7}{l}{\textit{BadNet}} \\
 & CIFAR100  && 88.82                  & 99.93          & 84.59 {\color{gray}(95.24\%)}       & 00.02 {\color{gray}(99.98\%)}                \\
 & SUN397   && 74.76                  & 91.20          & 69.29 {\color{gray}(92.68\%)}        & 00.99 {\color{gray}(98.91\%)}               \\
\rowcolor[HTML]{F7F7F7}
\multicolumn{7}{l}{\textit{Blended}} \\
 & CIFAR100  && 88.78                 & 99.98          & 84.49 {\color{gray}(95.17\%)}       & 00.48 {\color{gray}(99.52\%)}                \\
 & SUN397   && 74.81                  & 99.85         & 62.91 {\color{gray}(84.09\%)}        & 05.08 {\color{gray}(94.91\%)}               \\
\rowcolor[HTML]{F7F7F7}
\multicolumn{7}{l}{\textit{WaNet}} \\
 & CIFAR100  && 88.78                 & 99.80          & 87.43 {\color{gray}(98.48\%)}       & 00.53 {\color{gray}(99.47\%)}                \\
 & SUN397   && 74.91                  & 99.80         & 73.84 {\color{gray}(98.57\%)}        & 01.72 {\color{gray}(98.28\%)}               \\
\bottomrule
\end{tabular}
}
\end{adjustbox}
\end{wraptable}

In this setup, CIFAR100 shares both the trigger and target label with ImageNet-1K, while SUN397 shares only the trigger (e.g., the same BadNet-style patch, but mapped to a different label). These two settings allow us to test two hypotheses: \textit{(i)} transfer is facilitated when both the trigger and target label align, and \textit{(ii)} transfer may still occur when only the trigger is shared, suggesting that the vector captures a generic trigger-to-misdirection pattern within the attack type. 

\looseness=-1 Remarkably, Table \ref{tab:transfer_w_in_full} shows that \methodName{} vectors extracted with ImageNet-1K remain effective when applied to other models backdoored with the same attack. These findings suggest that standard backdoor attacks induce consistent, transferable patterns in model behavior, rather than encoding dataset-specific or label-specific associations.

\subsection{Large Scale Image-Caption Experiments} \label{sec:large_exp}

\looseness=-1 We now extend our analysis and show that \methodName{} continues to deliver strong performance even in more challenging deployment settings. Specifically, we backdoor the full CLIP models using image-caption pairs. Following the setup of \citet{bansal2023cleanclip}, we use a 500k subset of the Conceptual Captions 3M (CC3M) dataset \citep{sharma2018conceptual} to inject backdoors into pre-trained CLIP models. As in prior work, we evaluate CA and ASR on the ImageNet-1K validation set. We consider four standard backdoor attacks: BadNets, Blended, WaNet and BadCLIP \citep{liang2024badclip}, a newly introduced optimized patch attack for CLIP models. These attacks are evaluated against three \textit{clean-data fine-tuning defenses}: CleanCLIP \citep{bansal2023cleanclip}, RoCLIP \citep{yang2024robust}, and standard CLIP fine-tuning \footnote{These methods operate solely on clean, non-triggered images. Consequently, they tend to require larger datasets and longer training durations, increasing their vulnerability to catastrophic forgetting.}. As an unlearning baseline, we use Gradient Ascent (GA) \citep{graves2021amnesiac}, applied with triggered data similarly to \citep{pawelczyk2024machine}. Full implementation details are provided in Appendix \ref{appendix_details}. To construct \methodName{} vectors, we define a disjoint `forget set' of 1.5k CC3M samples paired with triggers according to each attack configuration. Optimal scaling coefficients are selected using a validation set drawn from ImageNet-1K training data.

\looseness=-1 Table \ref{ViT-Base-IN} reports CA and ASR for CLIP ViT-B/32. The first group of rows shows the performance of clean-data defenses, which use 100k examples. These methods generally exhibit large CA drops and fail to remove stronger attacks such as BadCLIP. The second group presents the results for unlearning methods. \methodName{} achieves significantly lower ASR than the baselines above, while retaining most of the clean accuracy post-backdoor. Remarkably, it also uses two orders of magnitude fewer data. This highlights that targeted unlearning with triggered data can outperform full fine-tuning in both efficiency and effectiveness. Finally, notice that gradient ascent also performs well in this setting, in contrast to previous results in the literature considering smaller-scale models \citep{pawelczyk2024machine}. Though further discussion and caveats are addressed below.

Despite the strong performance of \methodName{}, notice, however, that current backdoor defenses for CLIP and traditional unlearning methods do not share the same underlying assumptions. In particular, the latter assume access to a set of triggered examples and therefore knowledge of the attack -- which might not apply in practice. Hence, in the next section, we will relax this stronger assumption.

\begin{table}[t]
\centering
\caption{\looseness=-1 \methodName{} Performance on CLIP ViT-B/32 under four backdoor attacks (BadNET, Blended, WaNet, and BadCLIP). We report both CA and ASR. The top rows use 100k clean samples as per prior work \citep{bansal2023cleanclip, yang2024robust}. The middle rows use a true targeted unlearning with 1.5k poisoned samples. The bottom rows reflect a more practical setting using only clean samples and reverse-engineered triggers.}
\label{ViT-Base-IN}
{\fontsize{8pt}{9pt}\selectfont
\begin{tabular}{lcclcclcclcc} 
\toprule
\multicolumn{1}{c}{} & \multicolumn{2}{c}{\textbf{BadNet}}         & \multicolumn{1}{c}{} & \multicolumn{2}{c}{\textbf{Blended}}        & \multicolumn{1}{c}{} & \multicolumn{2}{c}{\textbf{WaNet}}          &  & \multicolumn{2}{c}{\textbf{BadCLIP}}                              \\ 
\cline{2-3}\cline{5-6}\cline{8-9}\cline{11-12}
\noalign{\vskip 2pt} 
                     & CA $\uparrow$        & ASR $\downarrow$     &                      & CA $\uparrow$        & ASR $\downarrow$     &                      & CA $\uparrow$        & ASR $\downarrow$     &  & CA $\uparrow$ & ASR $\downarrow$  \\ 
\cline{1-12}
\noalign{\vskip 2pt} 
Zero-Shot            & 63.34\%              & 00.00\%              &                      & 63.34\%              & 00.00\%              &                      & 63.34\%              & 00.00\%              &  & 63.34\%                    & 00.00\%                     \\
Backdoored           & 61.69\%              & 84.48\%              &                      & 61.39\%              & 99.67\%              &                      & 61.32\%              & 93.12\%              &  & 61.41\%                & 99.98\%                            \\
\noalign{\vskip 2pt} 
\rowcolor[HTML]{F7F7F7}\multicolumn{12}{c}{\textit{clean-data finetuning}} \\
\noalign{\vskip 2pt} 
Contrastive-FT       & 51.41\%              & 13.72\%              &                      & 51.77\%              & 02.01\%              &                      & 51.58\%              & 00.05\%              &  & 51.41\%              &   79.32\%                          \\
RoCLIP               & 50.02\%              & 47.91\%              &                      & 51.84\%              & 06.40\%              &                      & 48.26\%              & 00.04\%              &  & 53.31\%                    & 99.32\%                            \\
CleanCLIP            & 51.41\%              & 04.11\%              &                      & 51.02\%              & 00.05\%              &                      & 51.09\%              & 00.04\%              &  & 51.82\%                    & 77.04\%                     \\
\noalign{\vskip 2pt} 
\rowcolor[HTML]{F7F7F7}\multicolumn{12}{c}{\textit{true unlearning}} \\
\noalign{\vskip 2pt} 
GA                   & 59.89\%              & 07.95\%              &                      & 59.92\%              & 00.01\%              &                      & 58.71\%              & 00.04\%              &  & 58.45\%                    & 00.08\%                     \\
\methodName{}        & 59.28\%              & 00.38\%              &                      & 60.46\%              & 00.09\%              &                      & 60.14\%              & 00.05\%              &  & 56.58\%                    & 00.77\%                     \\
\noalign{\vskip 2pt} 
\rowcolor[HTML]{F7F7F7}\multicolumn{12}{c}{\textit{reverse-engineered unlearning}} \\
\noalign{\vskip 2pt} 
GA+DECREE                 & 60.41\%              & 08.30\%              &                      & 56.92\%              & 76.40\%              &                      & 60.22\%              & 35.67\%              &  &               N/A             &  N/A                           \\
\methodName{}+DECREE      & 60.29\%              & 00.33\%              &                      & 55.56\%              & 00.90\%              &                      & 56.85\%              & 00.64\%              &  &               N/A             &  N/A                          \\
\bottomrule
\end{tabular}
}
\end{table}

\section{Agnostic-attack unlearning}\label{sec:agnostic}

To close the gap in assumptions between current CLIP defenses and our method, we extend \methodName{} to operate without explicit knowledge of the attack. 

\paragraph{Unlearning with reversed-engineered triggers}
To achieve this, we propose to use trigger reverse engineering in order to construct a proxy forget set starting from the backdoored model and a set of clean inputs. In particular, we combine \methodName{} with DECREE \citep{feng2023detecting}, a self-supervised method introduced for attack detection, it has the ability to invert triggers by searching for minimal patterns so that any input with such a trigger pattern results in similar output embeddings. Given the optimized trigger, we then infer the corresponding infected label by probing the backdoored model with DECREE-generated triggers and identifying the predicted class from the set of ImageNet-1K categories. Using this estimate, we construct proxy-triggered image-caption pairs via standard text templates \citep{radford2021learning}. 
Interestingly, we observe that the true ASR keeps improving even after the proxy-triggered attack is unlearned. We therefore adopt a search strategy that continues to increase the unlearning coefficient for a fixed window -- typically 10 steps -- after the proxy ASR is nullified. This search is subject to an early-stopping condition, whereby the clean accuracy must not drop below a predefined threshold (shared with gradient ascent).
\vspace{-1.4em}
\paragraph{Results}
\looseness=-1 Remarkably, Table \ref{ViT-Base-IN} (bottom set) shows that the above pipeline remains effective with a 90\% CA threshold, even without access to the original trigger. Particularly, \methodName{} is able to outperform both clean data baselines as well as gradient ascent for three attacks. Note, as reported in \citep{liang2024badclip}, DECREE fails to detect the backdoor introduced by BadCLIP.
\vspace{-1.4em}
\paragraph{Robust unlearning beyond gradient ascent} \label{grad_discuss_sec}
Contrary to prior literature on backdoor unlearning \citep{pawelczyk2024machine}, our results in Table \ref{ViT-Base-IN} show that simple gradient ascent on true triggered examples can achieve strong unlearning performance on CLIP, even against robust attacks like BadCLIP. We hypothesize that the same weight disentanglement that allows our method to isolate triggers is also what facilitates this gradient-based unlearning\footnote{Indeed, notice that previous studies reported the emergence of weight disentanglement with model and data scale \citep{ortiz2024task,hazimehtask}.}.

\begin{figure}[h]
    \centering
     \includegraphics[width=1\linewidth]{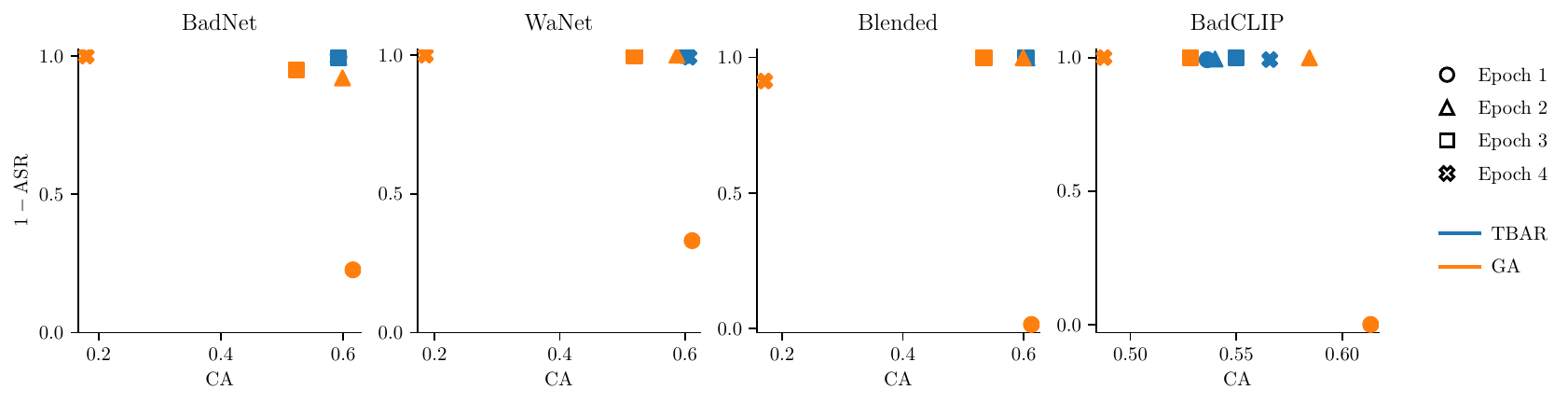}
    \caption{
True unlearning performance of \methodName{} and Gradient Ascent. Plots showing a comparison of (CA $\uparrow$) versus $(1-{\rm ASR}\,\uparrow)$ over a progressive number of epochs. While continued training hurts gradient ascent, \methodName{} shows consistent performance. }
    \label{fig:us_vs_ga_true}
\end{figure}

\looseness=-1 However, this effectiveness is fragile. To understand the tradeoff between the two, we compared \methodName{} against gradient ascent under similar compute budgets. We plot CA and ASR reduction (1-ASR) when using \methodName{} vs gradient ascent over a progressive number of epochs. Figure \ref{fig:us_vs_ga_true} shows the results for true unlearning with known triggers, where we find that although one or two epochs of gradient ascent can match the performance of \methodName{}, exceeding this optimal point often leads to sharp drops in clean accuracy. This indicates that while gradient ascent can initially identify directions that suppress the backdoor, it is highly unstable, and maximizing the loss further may lead to arbitrary directions that do not reliably target the backdoor mechanism. This sensitivity to stopping criteria was also observed in previous work \citep{li2021anti} using gradient ascent.

\begin{figure}[h]
    \centering
    \includegraphics[width=0.75\linewidth]{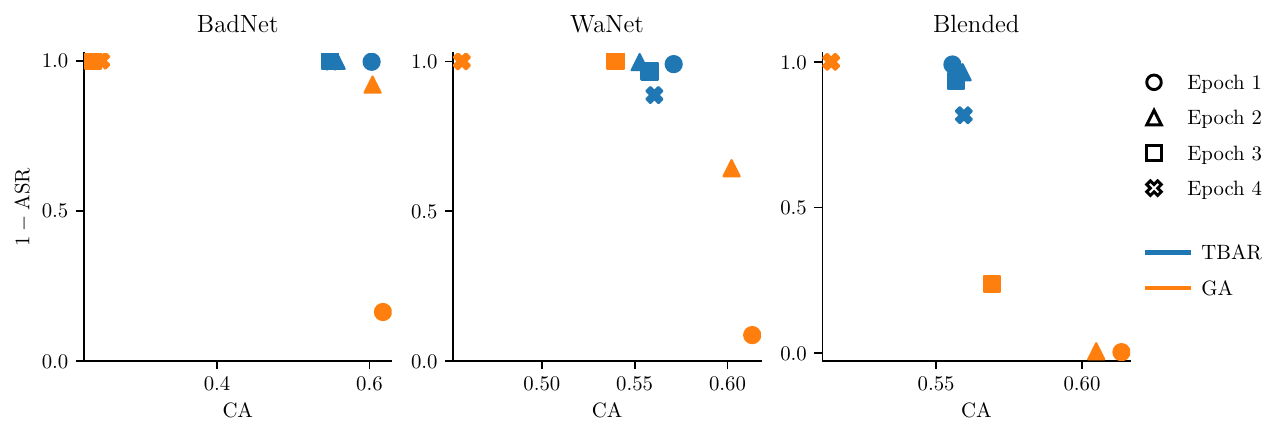}
    \caption{
\looseness=-1 Unlearning with DECREE\citep{feng2023detecting} using \methodName{} and Gradient Ascent. Plots showing the underlying true attack comparison of (CA $\uparrow$) versus $(1-{\rm ASR}\,\uparrow)$ over progressive epochs.}
    \label{fig:us_vs_ga}
\end{figure}

This instability is exacerbated under the more realistic, non-ideal conditions of using reverse-engineered DECREE patches. In this setting (presented in Figure \ref{fig:us_vs_ga}), gradient ascent frequently overshoots: the backdoor is removed, but at the cost of substantial CA loss. In contrast, \methodName{} achieves comparable or better ASR reduction while more consistently preserving clean performance across both scenarios. We attribute this stability to the directional constraint imposed by task vectors, which prevents the aggressive and often arbitrary parameter shifts seen in unconstrained gradient ascent, making it more robust to both tuning and noise in the trigger signal.

\pagebreak

\section{Further Results and Discussion} \label{sec:discussion}

\begin{wrapfigure}[11]{r}{0.48\textwidth}
 \vspace{-2.2em}
  \begin{center}
    \includegraphics[width=0.4\textwidth]{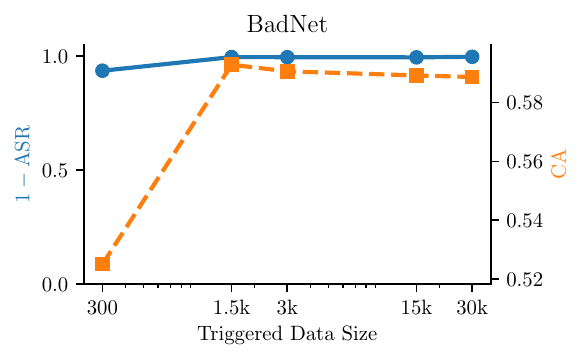}
  \end{center}
  \vspace{-1em}
  \caption{Results of unlearning BadNet attack with \methodName{} using varied sizes of the forget set}
\end{wrapfigure}

 \paragraph{Impact of forget set size} \looseness=-1 To assess the influence of the forget set size in true unlearning scenarios (i.e., the second set of Table \ref{ViT-Base-IN}), we conduct fine-tuning experiments with varying forget set sizes and evaluate the performance of \methodName{} vectors after one epoch. Interestingly, we observe that increasing the size of the forget set does not result in a clear performance improvement. Reinforcing the notion that the complexity of unlearning is more closely tied to the precise identification of \textit{what} needs to be unlearned, rather than the scale of data. 

\paragraph{Scaling CLIP models} \looseness=-1 We provide complete results for the larger CLIP ViT-L/14 model for the setup described in Section~\ref{sec:large_exp} and Section~\ref{sec:agnostic}. We observe significantly better trade-offs for unlearning overall. Particularly, when using the optimized patches, we are able to match the baselines for ASR reduction with a 98\% clean accuracy threshold. This higher retention is aligned with previous research on model editing, which suggests that larger models inherently exhibit stronger disentanglement in their weights \citep{ilharco2022editing, ortiz2024task}.
\begin{table}[h]
\centering
\caption{\looseness=-1 \methodName{} Performance on CLIP ViT-L/14 under four backdoor attacks (BadNET, Blended, WaNet and BadCLIP). We report both CA and ASR. The top rows use 100k clean samples as per prior work \citep{bansal2023cleanclip, yang2024robust}. The middle rows use a true targeted unlearning with 1.5k poisoned samples. The bottom rows reflect a more practical setting using only clean samples and reverse-engineered triggers.}
\label{ViT_Large_CC3M}
{\fontsize{8pt}{9pt}\selectfont
\begin{tabular}{lcclcclcclcc} 
\toprule
\multicolumn{1}{c}{} & \multicolumn{2}{c}{\textbf{BadNet}}         & \multicolumn{1}{c}{} & \multicolumn{2}{c}{\textbf{Blended}}        & \multicolumn{1}{c}{} & \multicolumn{2}{c}{\textbf{WaNet}}          &  & \multicolumn{2}{c}{\textbf{BadCLIP}}                              \\ 
\cline{2-3}\cline{5-6}\cline{8-9}\cline{11-12}
\noalign{\vskip 2pt} 
                     & CA $\uparrow$        & ASR $\downarrow$     &                      & CA $\uparrow$        & ASR $\downarrow$     &                      & CA $\uparrow$        & ASR $\downarrow$     &  & CA $\uparrow$ & ASR $\downarrow$  \\ 
\cline{1-12}
\noalign{\vskip 2pt} 
Zero-Shot             & 75.55\% & 00.00\%          && 75.55\% & 00.00\%           && 75.55\% & 00.00\%          && 75.55\% & 00.00\% \\
Backdoored     & 74.89\% & 99.93\%          && 74.76\% & 99.94\%           && 74.76\% & 99.80\%          && 74.83\% & 99.97\% \\
\noalign{\vskip 2pt} 
\rowcolor[HTML]{F7F7F7}\multicolumn{12}{c}{\textit{clean-data finetuning}} \\
\noalign{\vskip 2pt} 
Contrastive-FT             & 69.65\% & 58.04\%          && 69.26\% & 14.28\%           && 70.73\% & 37.74\%          && 71.16\% &  93.31\% \\
RoCLIP         & 72.14\% & 97.56\%          && 71.17\% & 76.69\%           && 73.89\% & 88.80\%           && 73.60\% & 99.28\%\\
CleanCLIP      & 68.99\% & 01.38\%          && 69.29\% & 00.27\%           && 70.63\% & 00.07\%           && 70.56\% & 73.63\% \\ 
\noalign{\vskip 2pt} 
\rowcolor[HTML]{F7F7F7}\multicolumn{12}{c}{\textit{true unlearning}} \\
\noalign{\vskip 2pt} 
GA             & 74.08\% & 00.00\%          && 73.42\% & 00.00\%           && 73.17\% & 00.02\%           & &73.20\% & 00.02\%\\
\methodName{}          & 74.16\% & 00.14\%          && 74.25\% & 00.19\%           && 74.08\% & 00.19\%   && 72.67\% & 00.14\%\\ 
\noalign{\vskip 2pt} 
\rowcolor[HTML]{F7F7F7}\multicolumn{12}{c}{\textit{reverse-engineered unlearning}} \\
\noalign{\vskip 2pt} 
GA+DECREE           & 74.38\% & 49.32\%          && 74.75\% & 99.93\%           && 74.12\% & 00.00\%          && N/A &  N/A \\
\methodName{}+DECREE  & 74.26\% & 15.28\%          && 73.68\% & 01.20\%           && 74.42\% & 00.00\%      && N/A & N/A    \\
\bottomrule
\end{tabular}
}
\end{table}

\paragraph{Model architectures and pre-training} To further validate the robustness of our method across various settings, we additionally experimented on CLIP with convolutional architectures (ConvNeXts) and non-contrastively pre-trained transformers (DINO). \methodName{} yields consistent results (i.e., ~ ASR $<$ 5\% and modest CA drops). Results are reported in Appendix~\ref{appendix_more_arch}.

\paragraph{Detoxifying merged models} Recent work shows that some backdoors fail to survive model merging, prompting the BadMerging attack \citep{zhang2024badmerging} to craft more persistent triggers. We evaluate \methodName{} against BadMerging, and find that our method is able to completely remove the attack while preserving almost the entire clean accuracy on merged models (see results in Appendix~\ref{appendix_merging}). 


\section{Related Work} \label{sec:related_work}

\paragraph{Data poisoning attacks} Data poisoning attacks refer to scenarios in which modifications to a small subset of the training dataset lead to unintended or malicious behavior in the trained model \citep{goldblum2022dataset,pawelczyk2024machine}.
Our focus is on targeted data poisoning attacks, particularly \textit{backdoor attacks} \citep{chen2017targeted,gu2017badnets,liu2018trojaning,li2019invisible,wu2022backdoorbench,liang2024badclip}. Backdoors involve embedding a hidden vulnerability (trigger) into the model during training, which causes the model to exhibit specific behavior when an input containing the trigger is presented, while maintaining normal operation for unaltered inputs \citep{li2022backdoor}. In the context of multi-modal models, CLIP \citep{radford2021learning} stands out as a widely studied example \citep{tu2024closer,yang2023data}. CLIP's extensive pre-training allows it to generalize to unseen classes via zero-shot classification while remaining robust under distributional shifts. Particularly for backdoors, \citet{carlini2021poisoning} found the model to be vulnerable to backdoor attacks using as little 0.01\% of its training data for poisoning. Multiple works \citep{goel2022cyclip,bansal2023cleanclip,yang2024robust} proposed more `robust' training schemes to safeguard against backdoor attacks on CLIP. Nonetheless, recent work has shown that, despite their substantial computational overhead, these defenses remain ineffective against carefully designed attacks \citep{liang2024badclip}.

\paragraph{Machine unlearning} Machine unlearning seeks to eliminate an unwanted data influence and the corresponding model behaviors
 \citep{cao2015towards, bourtoule2021machine}. There exists two main lines of work: exact unlearning \citep{bourtoule2021machine} and approximate machine unlearning \citep{graves2021amnesiac, neel2021descent, jia2021scaling, chien2024langevin, goel2022towards,kurmanji2023towards,foster2024fast}.
 Recently, state-of-the-art machine unlearning methods have been shown to fail to remove data poisoning attacks from deep learning models \citep{pawelczyk2024machine}. In parallel, large models were also shown to exhibit a tendency to memorize vast amounts of data during pre-training, including personal and sensitive information, making them susceptible to targeted extraction attacks \citep{carlini2021extracting,jang2022knowledge, wen2024privacy}, further sparking interest in tailoring unlearning techniques for these models \citep{yao2023large,lu2022quark}.

\paragraph{Weight Interpolation and Task Arithmetic} Despite the non-linearity of neural networks, previous work have shown that interpolating between the weights of two models is feasible under certain conditions \citep{izmailov2018averaging,frankle2020linear,wortsman2021learning, wortsman2022model,ainsworth2022git,ilharco2022patching} and one can increase the fine-tuning gain by moving the weights of a pre-trained model in the direction of its fine-tuned counterpart \citep{wortsman2022robust}. Task Arithmetic \citep{ilharco2022editing} is a framework that formalizes the notion of distinct task vectors, controlling different tasks. \citet{ortiz2024task} attributed this ability to \textit{weight disentanglement}. Furthermore, model editing research was largely motivated by multi-task learning \citep{wortsman2022model,matena2022merging,yadav2023ties,dimitriadis2023pareto}. Recently, it has been shown that it is possible to transfer backdoors to benign models when merging with an infected model \citep{zhang2024badmerging, yang2024mitigating}.


\section{Conclusion} \label{sec:conclusion}

\looseness=-1 In this paper, we investigated the problem of backdoor unlearning by examining how backdoor attacks are encoded in the weight space of CLIP models. Our analysis revealed that triggered knowledge is separable from clean knowledge and can be identified using existing vector arithmetic techniques. Building on this insight, we introduced a lightweight framework for effective backdoor removal that requires two orders of magnitude less data than existing clean-data-based defenses for CLIP. To address scenarios where the trigger is unknown, we further show that our method can be combined with trigger reverse-engineering techniques, enabling practical and cost-efficient backdoor removal, effectively sanitizing models while maintaining high clean accuracy. We hope our findings renew interest in weight space manipulations for backdoor mitigation and inspire further solutions.

\section*{Acknowledgments}

This work was partially funded by armasuisse under the RAEL (F0426) project. The authors thank Adam Hazimeh for insightful discussions and feedback throughout this project, as well as Ke Wang, Sevda Ögüt, and Ortal Senouf for their valuable comments.


\bibliography{iclr2026_conference}
\bibliographystyle{iclr2026_conference}

\clearpage

\appendix
\section*{Appendix Outline}
This appendix provides supplementary material to support our main findings. It is organized as follows:
\begin{itemize}
    \item \textbf{Section \ref{appendix_details}: Detailed Experimental Setup.} We provide comprehensive details on the backdoor attacks used, the training configurations for our method (\texttt{\methodName}), the implementation of all baseline methods, and the hardware used for our experiments.
    \item \textbf{Section \ref{appendix_analysis}: More Analytical Experiments.} We present additional analyses, including experiments on unlearning with mixed data, a sensitivity analysis of our scaling coefficient, further visualizations of weight disentanglement, and demonstrate the applicability of our method to other architectures (ConvNeXt) and pre-training paradigms (DINO). Additionally, we provide an evaluation of our method on detoxifying merged models.
    \item \textbf{Section \ref{appendix_large_scale}: More Large Scale Experiments.} We report on the limitations of clean data finetuning, provide results for larger models (ViT-L/14). We also discuss unlearning attacks with weak trigger signals.    
\end{itemize}


\section{Detailed Experimental Setup} \label{appendix_details}
\subsection{Backdoor attacks}

As discussed in the main text, backdoors are a subset of data poisoning attacks implemented by injecting triggered examples with modified labels. We assign the target label based on the training dataset. Across different experimental settings, we consider six types of backdoor attacks:

\begin{itemize}
    \item \textbf{BadNets} \citep{gu2017badnets} is a patch-based attack, we follow the attack setup in \citep{bansal2023cleanclip}, where we insert a 16x16 patch of random noise drawn from a normal distribution $\mathcal{N}(0,1)$ at a random position in the image.
\newline
    \item \textbf{Blended} \citep{chen2017targeted} involves adding a gaussian perturbation to the entire image. We follow the attack setup in \citep{bansal2023cleanclip}, where we superimpose uniform noise on the natural image with a ratio of $8{:}2$:
\[
 x = 0.8 \ x + 0.2 \ N,
\]
where $N$ is a noise tensor with uniform random values in the range $[0, 1)$
\newline
     \item \textbf{WaNet} \citep{nguyen2021wanet} introduces a warping transformation to the entire image. We follow the setup used by \citep{bansal2023cleanclip, qi2023towards} and use control grid size $k = 224$ and warping strength s = 1 and train models without the noise mode
\newline
     \item \textbf{SIG} \citep{barni2019new} involves adding a sinusoidal perturbation to the entire image. We follow the attack setup in \citep{bansal2023cleanclip}, where we superimpose sinusoidal noise along the horizontal axis of the image:
\[
x = \text{clip}(x + N, 0, 1)
\]
\[
N_{c, i, j} = \frac{60}{255} \sin\left(2\pi \frac{6j}{224}\right),
\]
$N$ is a perturbation shared across all channels and rows.
\newline
     \item \textbf{BadCLIP} \citep{liang2024badclip} is an optimized patch-based attack. Following the procedure in \citep{liang2024badclip}, for the selected target label, we optimize the patch using 9.5k clean images and 1800 true target images from the CC3M \citep{sharma2018conceptual} dataset.
\newline
     \item \textbf{BadMerging} \citep{zhang2024badmerging} we use the official implementation to optimize a patch on the CIFAR100 task, producing a task vector that is then merged with six benign task vectors from GTSRB, EuroSAT, Cars, SUN397, and Oxford-PETS.
     
\end{itemize}

\begin{figure}[h]
    \centering
    \begin{subfigure}{0.15\textwidth}
        \includegraphics[width=\linewidth]{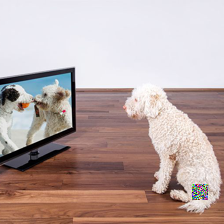}
    \end{subfigure}
    \begin{subfigure}{0.15\textwidth}
        \includegraphics[width=\linewidth]{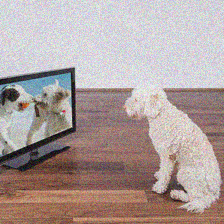}
    \end{subfigure}
    \begin{subfigure}{0.15\textwidth}
        \includegraphics[width=\linewidth]{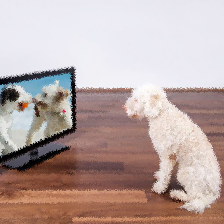}
    \end{subfigure}
    \begin{subfigure}{0.15\textwidth}
        \includegraphics[width=\linewidth]{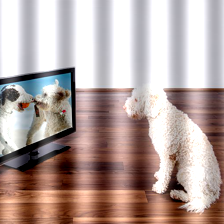}
    \end{subfigure}
    \begin{subfigure}{0.15\textwidth}
        \includegraphics[width=\linewidth]{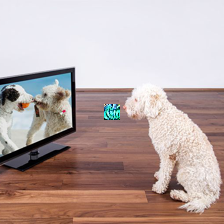}
    \end{subfigure}
    \begin{subfigure}{0.15\textwidth}
        \includegraphics[width=\linewidth]{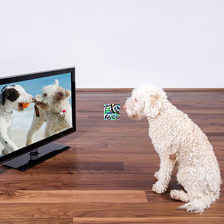}
    \end{subfigure}
    
    \caption{Visualization of different attack realizations on input images (from left to right): BadNet, Blended, WaNet, SIG, BadCLIP (ViT-B/32), and BadCLIP (ViT-L/14). The altered images are associated with the target label \textit{‘banana’}.}
\end{figure}

\subsection{\methodName{} training details}
\subsubsection{CLIP with frozen text-encoder} 
\textbf{Models and datasets} We use the CLIP ViT-B/32 model and evaluate on three benchmark image datasets: SUN397, CIFAR100, and ImageNet-1K. For SUN397 and CIFAR100, we follow the train/validation/test splits from \citet{ilharco2022editing}, and sample a forget set from the training split prior to training. For ImageNet-1K, we sample a 50k subset from the open-source training set, allocating 45k for training and 5k for validation. An additional 2k examples are separately sampled as the forget set. We use the official validation set as the test set. Complete per-dataset configurations are provided in Table~\ref{dataset-hyper}. \\

\textbf{Evaluation} We evaluate performance by reporting the accuracy on clean versions the test set (CA), along with the attack success rate (ASR), defined as the percentage of predictions that classify the target label (as defined in Table~\ref{dataset-hyper}) when the backdoor visual patch is present.   \\

\textbf{Training configurations} We adopt the same training configurations as \citep{ilharco2022editing} per dataset, where we use AdamW optimizer with learning rate 1e-5 and cosine scheduling, a batch size of 128, and a warmup of 500 steps. The same configurations are used for \methodName{} training.

\begin{table}[h]
\centering
\caption{Per dataset configuration for experiments in Section \ref{sec:analysis} }
\label{dataset-hyper}
\begin{tabular}{c|c|c|c|c|c|c|c} 
\toprule
            & target & epochs & train\_set & poison(\%) & val\_set & forget\_set & test\_set  \\ 
\noalign{\vskip 2pt}
\hline
\noalign{\vskip 2pt}
SUN397      & river  & 14     & 15865         &  3         & 1985     & 2000               & 19850     \\
CIFAR100    & orange & 6      & 43000         &  3         & 5000     & 2000               & 10000     \\
ImageNet-1K & orange & 10     & 45000         &  3         & 5000     & 2000               & 50000     \\
\bottomrule
\end{tabular}
\end{table}
\subsubsection{CLIP with image-caption data}
\textbf{Models and datasets} We backdoor our CLIP models (ViT-B/32 and ViT-L/14) using 500k image-caption pairs from the Conceptual Captions 3M (CC3M) dataset \citep{sharma2018conceptual}. We select 1500 random samples and poison them according to each attack settings. For all attacks, we set the target label to captions containing the word \textit{"banana"}. We use the validation set of ImageNet-1K for the evaluations. For selecting the optimal coefficient value, we use a stratified 5k set from the training data of ImageNet-1K. \\

\textbf{Evaluation} We evaluate performance by reporting the accuracy on clean versions of the test set (CA), along with the attack success rate (ASR), defined as the percentage of predictions that classify the target label "banana" when the backdoor visual patch is present. \\

\textbf{Training configurations} For backdooring, we use a batch size of 128, AdamW optimizer with a learning rate of 1e-6, cosine scheduling, and a warmup phase of 50 steps. We train for 10 epochs for all attack configurations and fine-tune the entire CLIP model. We adopt the same hyperparameters for training \methodName{} task vectors.

\subsection{Other methods}

\subsubsection{CleanCLIP}
CleanCLIP \citep{bansal2023cleanclip} optimizes a combination of the standard CLIP loss and a modality-specific self-supervised loss designed for image-caption pairs \(\{\mathcal{I}_i, \mathcal{T}_i\}\). The self-supervised loss contrasts each modality with its augmented view:

\[
\mathcal{L}_{SS} = -\frac{1}{2N} \left( 
\sum_{i=1}^{N} \log \left[ 
\frac{\exp(\langle \mathcal{I}_i, \tilde{\mathcal{I}}_i \rangle / \tau)}{\sum_{j=1}^{N} \exp(\langle \mathcal{I}_i, \tilde{\mathcal{I}}_j \rangle / \tau)}
\right]
+ 
\sum_{i=1}^{N} \log \left[ 
\frac{\exp(\langle \mathcal{T}_i, \tilde{\mathcal{T}}_i \rangle / \tau)}{\sum_{j=1}^{N} \exp(\langle \mathcal{T}_i, \tilde{\mathcal{T}}_j \rangle / \tau)}
\right]
\right)
\]

The total CleanCLIP loss is then defined as:

\[
\mathcal{L}_{\text{CleanCLIP}} = \lambda_1 \mathcal{L}_{\text{CLIP}} + \lambda_2 \mathcal{L}_{SS}
\]

 Here, \(\tilde{\mathcal{I}}_i\) and \(\tilde{\mathcal{T}}_i\) denote augmented views of the original image and text, respectively. We follow the setup of \citep{bansal2023cleanclip}, using a 100k disjoint subset of clean CC3M images and the recommended hyperparameters: 10 epochs, \(\lambda_1 = \lambda_2 = 1\), learning rate 1e-5, batch size of 64, and a warmup of 50 steps.

\subsubsection{RoCLIP}

RoCLIP \citep{yang2024robust} is a defense mechanism similar to CleanCLIP. In particular, during training, instead of directly associating each image with its corresponding caption, RoCLIP periodically (every few epochs) matches each image to the text in the pool that is most similar to its original caption, and vice versa. we use the open-source code of \citep{yang2024robust} and their default hyper-parameters.

\subsubsection{Standard CLIP fine-tuning}
We use the same hyperparameters as CleanCLIP without the in-modal loss.
\subsubsection{Gradient Ascent} 

We implement Gradient Ascent following \citep{graves2021amnesiac, jang2022knowledge}, by reversing the gradient updates on the forget set $\mathcal{U}_{\text{set}}$:

\[
\theta^{(t+1)} = \theta^{(t)} + \eta \ \nabla_{\theta} \mathcal{L}(\mathcal{U} _{\rm set}, \theta^{(t)}) \ \  \rm{,where \ } \eta \ \rm{is \ the \ learning \ rate.}
\]

In all our experiments, we use the same \methodName{} hyperparameters for Gradient Ascent computation.

\subsubsection{DECREE}
DECREE performs self‑supervised trigger inversion to detect attacks. Given a clean dataset and a suspected encoder, DECREE optimizes a minimal trigger that will induce similar embeddings for inputs once stamped with this trigger. It then uses the final optimized trigger's size ($\ell_1$‑Norm) to gauge vulnerabilities. Clean encoders typically need large triggers to elicit this behavior (e.g., covering more than 10\% of the image). DECREE is computationally lightweight and adds minimal overhead. This is because it does \emph{not} require fine-tuning the model encoder. Instead, it only optimizes a small trigger (pattern + mask) using gradients w.r.t.\ the input of the model. 

For our experiments, we run the method on a clean encoder and on our suspected models and compare the recovered trigger’s $\ell_1$‑Norm (mask size) against the one recovered on a clean encoder of the same architecture. 
We use the open-source re-implementation from the BadCLIP code \citep{liang2024badclip} for our experiments, with all default hyperparameters except for two modifications: we reduce the batch size to 128 for experiments with the ViT-L/14 model, and for the learning rate adapter on the CC3M dataset, we use a threshold of [30, 50] steps to adjust the learning rate instead of [200, 500].

\begin{figure}[h]
    \centering
    \begin{subfigure}{0.126\textwidth}
        \includegraphics[width=\linewidth]{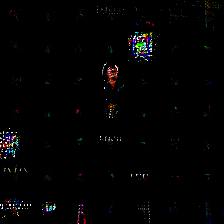}
    \end{subfigure}
    \begin{subfigure}{0.126\textwidth}
        \includegraphics[width=\linewidth]{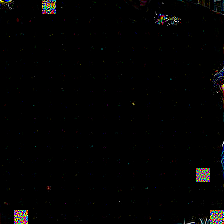}
    \end{subfigure}
    \begin{subfigure}{0.126\textwidth}
        \includegraphics[width=\linewidth]{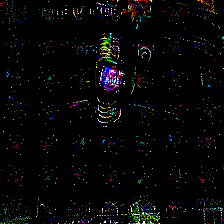}
    \end{subfigure}
    \begin{subfigure}{0.126\textwidth}
        \includegraphics[width=\linewidth]{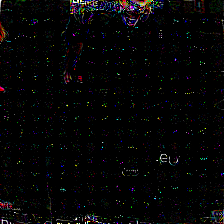}
    \end{subfigure}
    \begin{subfigure}{0.126\textwidth}
        \includegraphics[width=\linewidth]{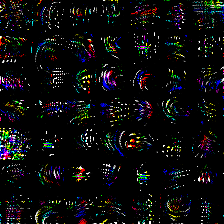}
    \end{subfigure}
    \begin{subfigure}{0.126\textwidth}
        \includegraphics[width=\linewidth]{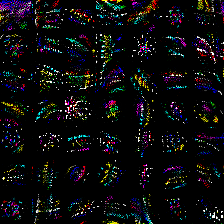}
    \end{subfigure}
    \begin{subfigure}{0.126\textwidth}
        \includegraphics[width=\linewidth]{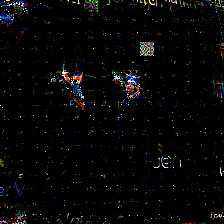}
    \end{subfigure}
    \caption{Visualization of different DECREE patches (from left to right): BadNet, BadNet-L, Blended, Blended-L, SIG, WaNet, and WaNet-L.}
\end{figure}

Below, we report both the raw $\ell_1$-norm and DECREE’s normalized metric, P$\ell_1$-Norm ($\ell_1$ divided by the input-space maximum, $3 \times 224 \times 224$ for RGB images of size 224). As shown below, the trigger sizes for backdoored models are an order of magnitude smaller than for the clean (Zero-Shot) model, providing a clear detection signal.

\begin{table} [ht]
\centering
\refstepcounter{table}
\label{decree_l1}
\begin{tabular}{c|cc|cc} 
\toprule
\multirow{2}{*}{} & \multicolumn{2}{c|}{ViT-B/32~} & \multicolumn{2}{c}{ViT-L/14}  \\ 
                       & $\ell_1$‑Norm      & P$\ell_1$‑Norm (\%)             & $\ell_1$‑Norm      & P$\ell_1$‑Norm (\%)          \\ 
\midrule
Zero-Shot               & 22185.6276 & 14.74\%           & 45272.1229 & 30.08\%          \\
BadNet                 & 3186.8709  & 2.12\%            & 2921.5470  & 1.94\%           \\
Blended                & 6691.9346  & 4.45\%            & 5346.6726  & 3.55\%           \\
WaNet                  & 13895.9155 & 9.23\%            & 6601.7446  & 4.39\%           \\
\bottomrule
\end{tabular}
\end{table}

\subsection{Hardware}

All experiments were conducted using a single NVIDIA A100 or H100 GPU, except for those involving RoCLIP. Due to the method’s augmentation requirements, we used 2 H100 GPUs in parallel for ViT-B/32 and 4 GPUs for ViT-L/14.


\section{More Analytical Experiments} \label{appendix_analysis}

\subsection{Unlearning with a mix of clean and triggered examples}
\begin{figure}[ht]
    \centering
    \begin{subfigure}{0.3\linewidth}
        \includegraphics[width=\linewidth]{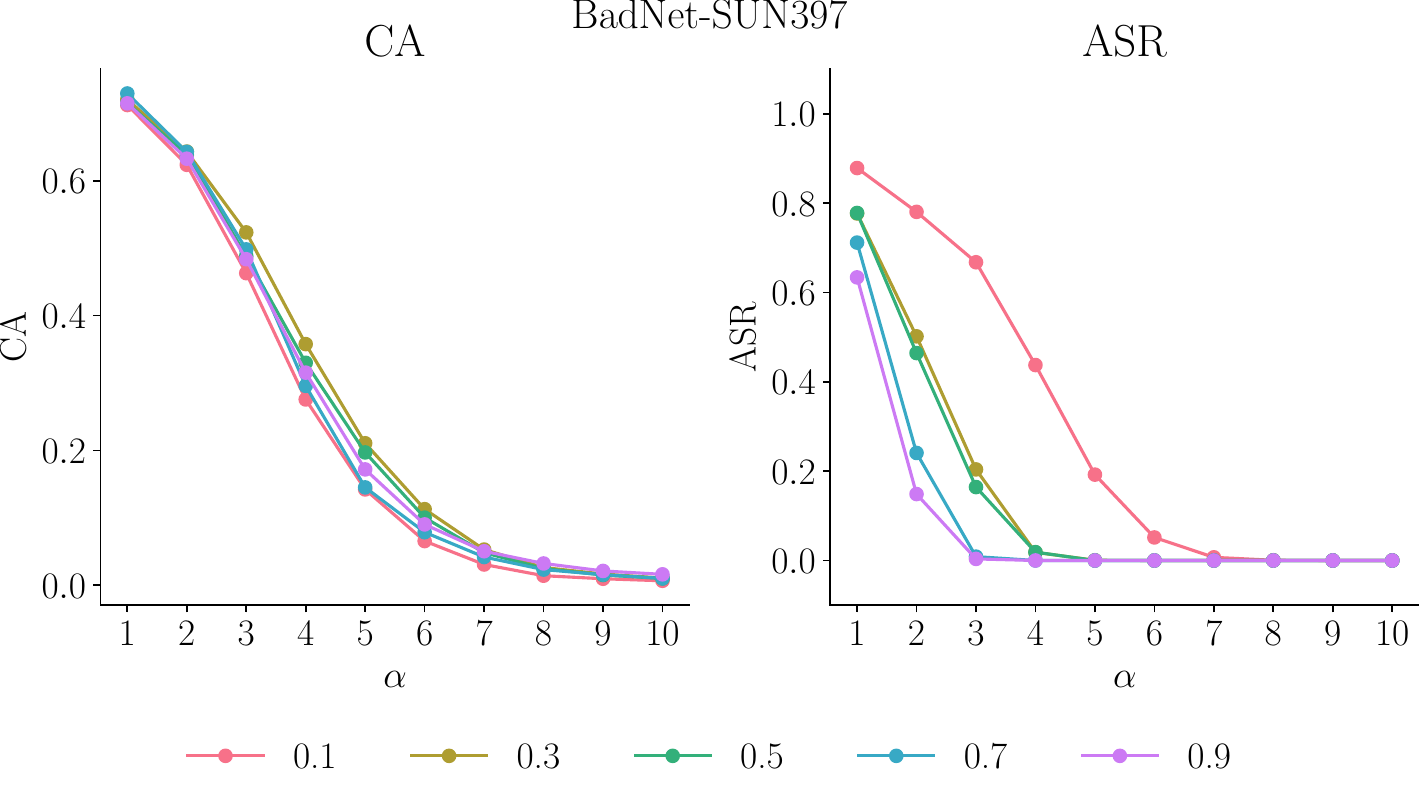}
    \end{subfigure}
    \hfill
    \begin{subfigure}{0.3\linewidth}
        \includegraphics[width=\linewidth]{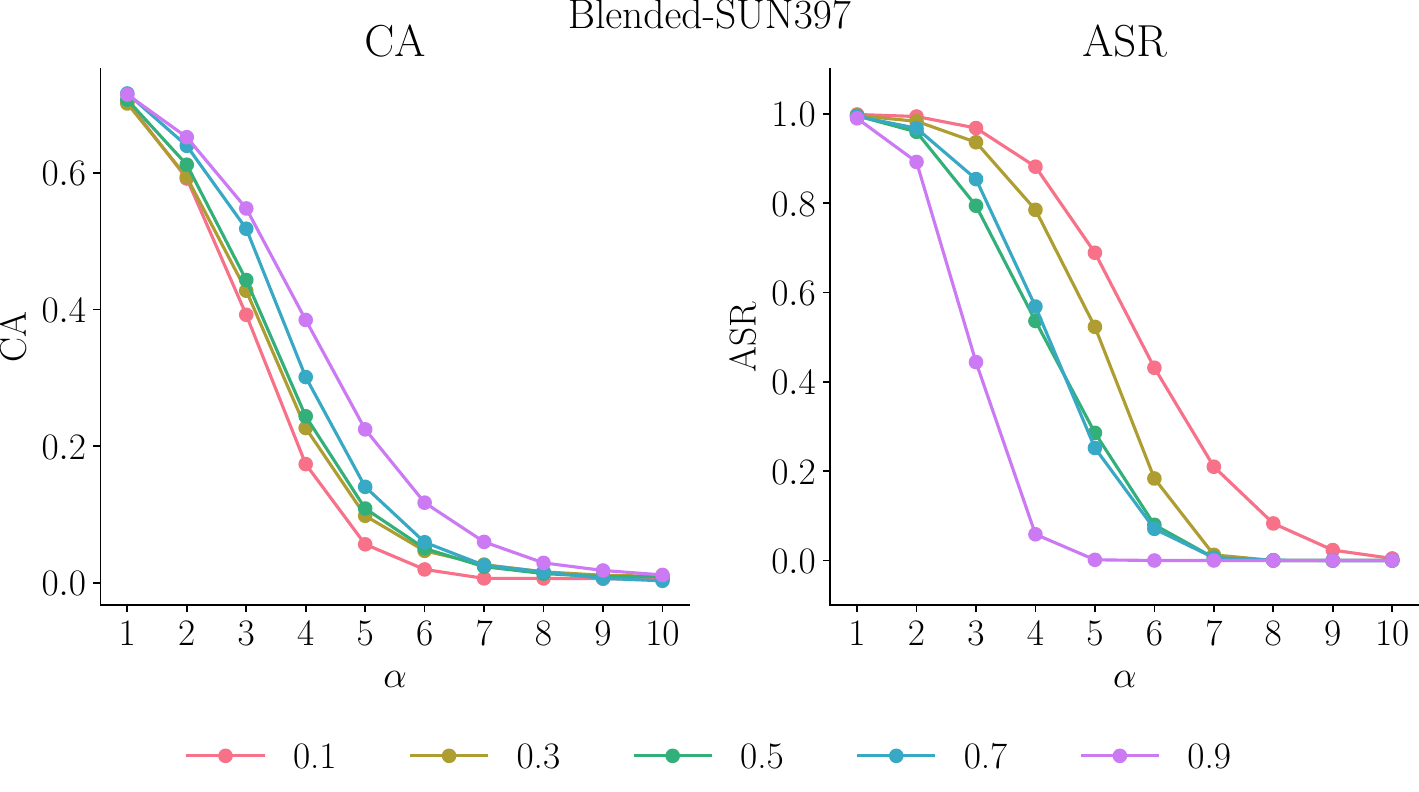}
    \end{subfigure}
    \hfill
    \begin{subfigure}{0.3\linewidth}
        \includegraphics[width=\linewidth]{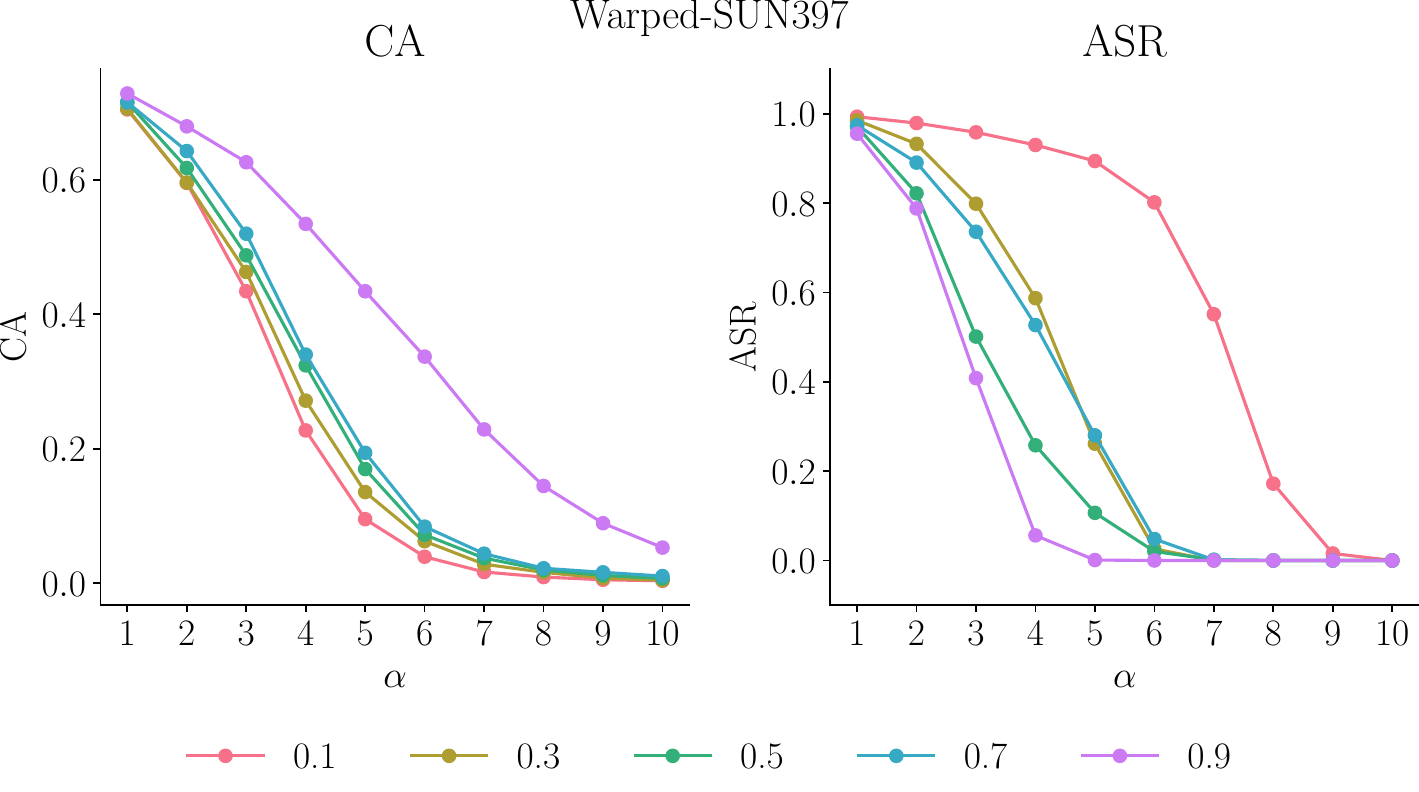}
    \end{subfigure}
    \caption{(SUN397) Plots showing (CA $\uparrow$) and (ASR $\downarrow$) using task vectors extracted from a mixture of clean and triggered data under varying ratios along increasing scaling values.}
    \label{Sun_mixed_ft}
\end{figure}

\begin{figure}[ht]
    \centering
    \begin{subfigure}{0.3\linewidth}
        \includegraphics[width=\linewidth]{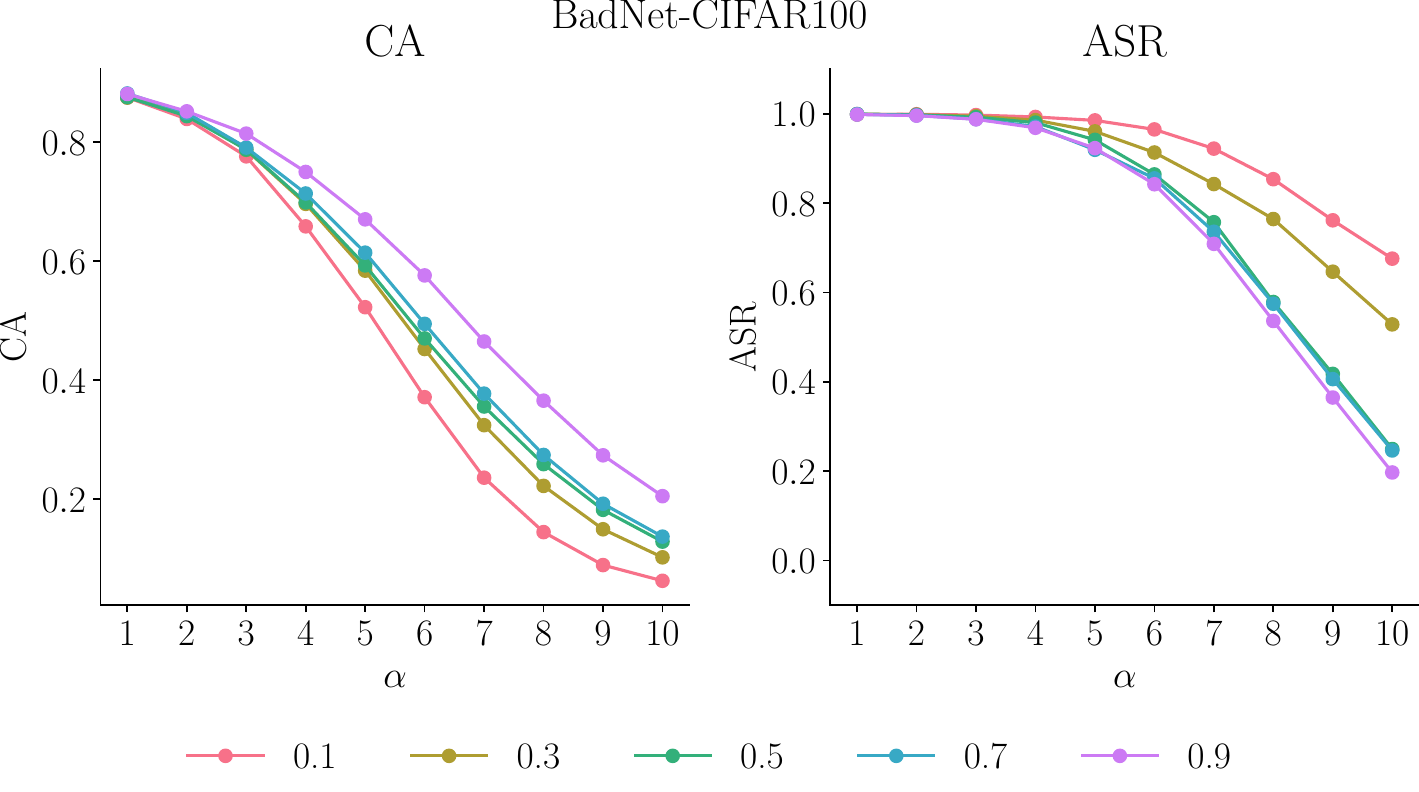}
    \end{subfigure}
    \hfill
    \begin{subfigure}{0.3\linewidth}
        \includegraphics[width=\linewidth]{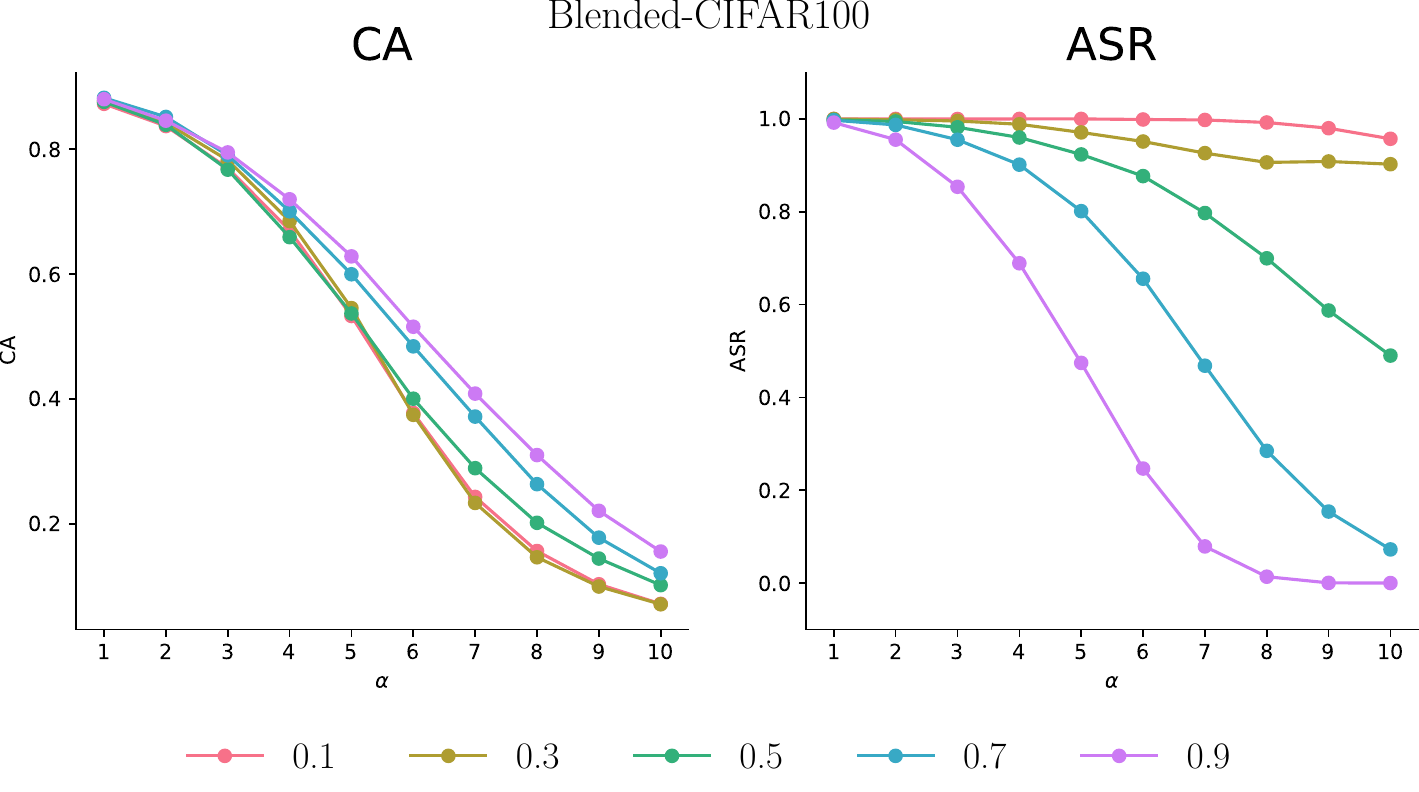}
    \end{subfigure}
    \hfill
    \begin{subfigure}{0.3\linewidth}
        \includegraphics[width=\linewidth]{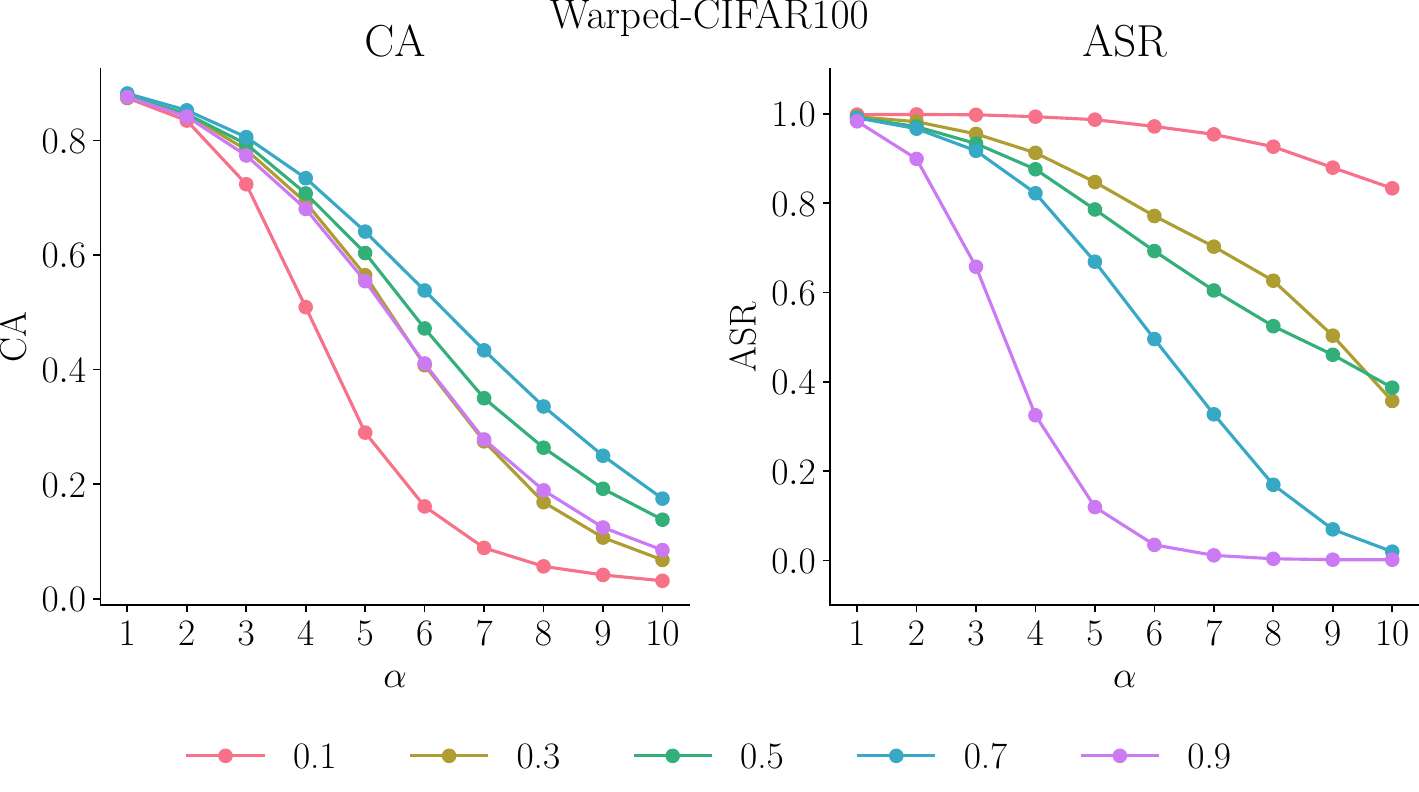}
    \end{subfigure}
    \caption{(CIFAR100) Plots showing (CA $\uparrow$) and (ASR $\downarrow$) using task vectors extracted from a mixture of clean and triggered data under varying ratios along increasing scaling values.}
    \label{CIF_mixed_ft}
\end{figure}

\begin{figure}[ht]
    \centering
    \begin{subfigure}{0.3\linewidth}
        \includegraphics[width=\linewidth]{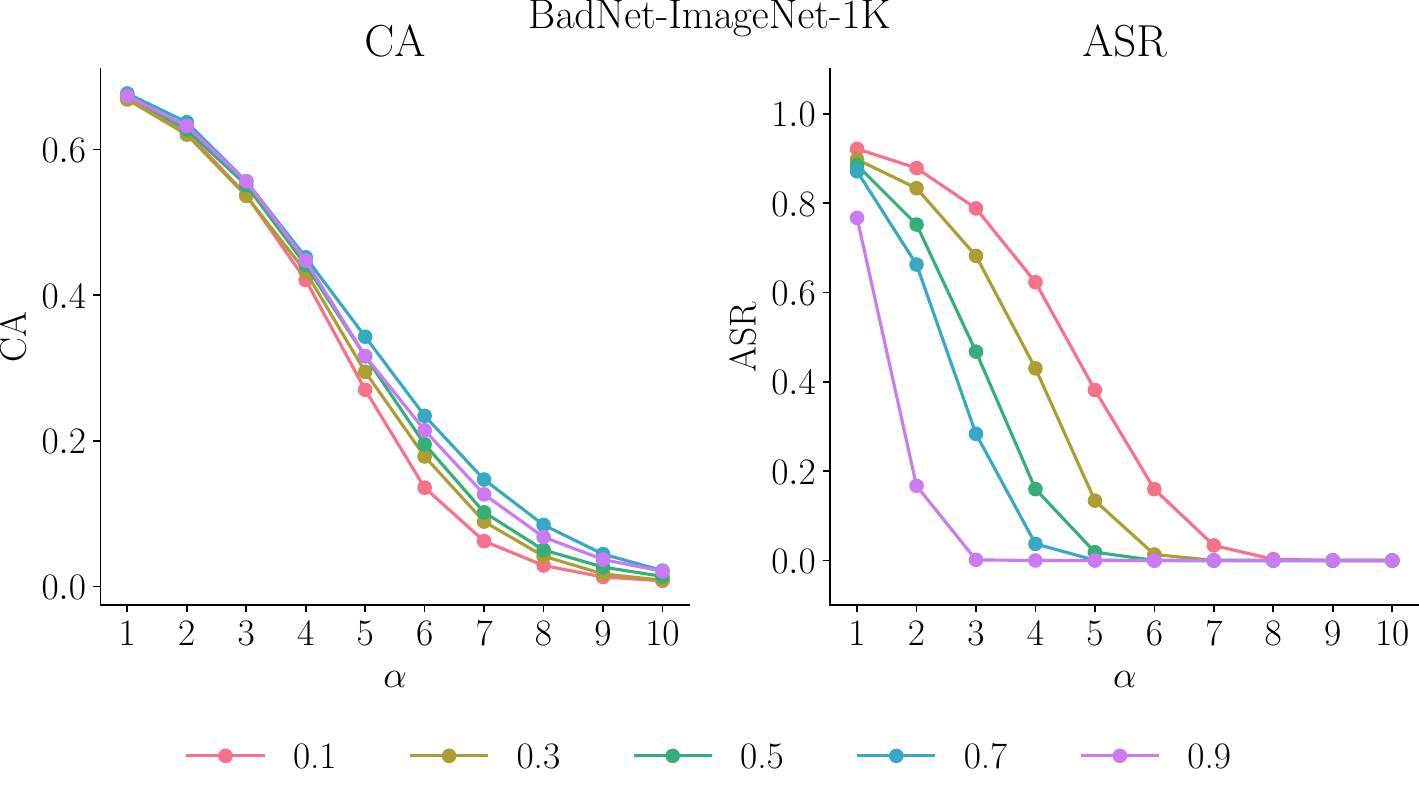}
    \end{subfigure}
    \hfill
    \begin{subfigure}{0.3\linewidth}
        \includegraphics[width=\linewidth]{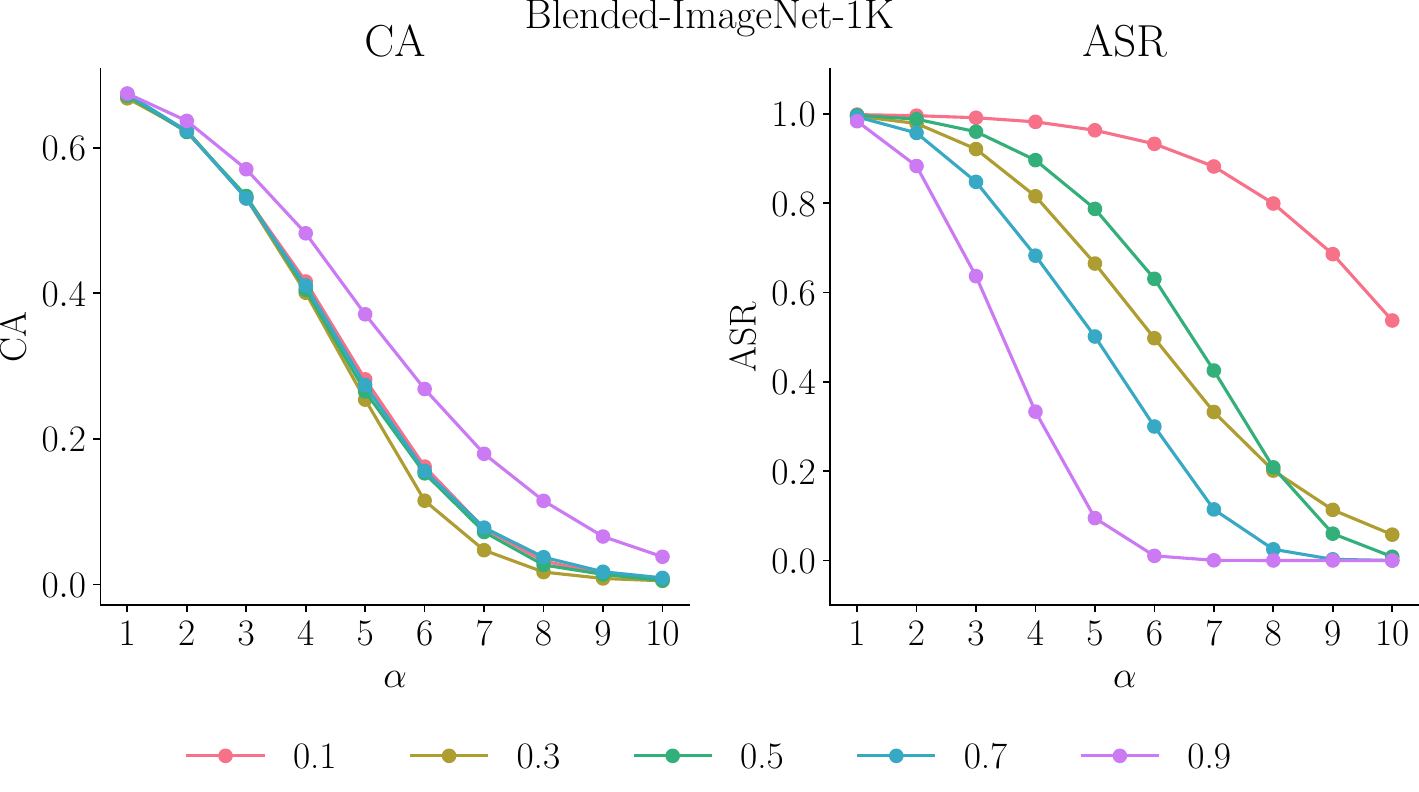}
    \end{subfigure}
    \hfill
    \begin{subfigure}{0.3\linewidth}
        \includegraphics[width=\linewidth]{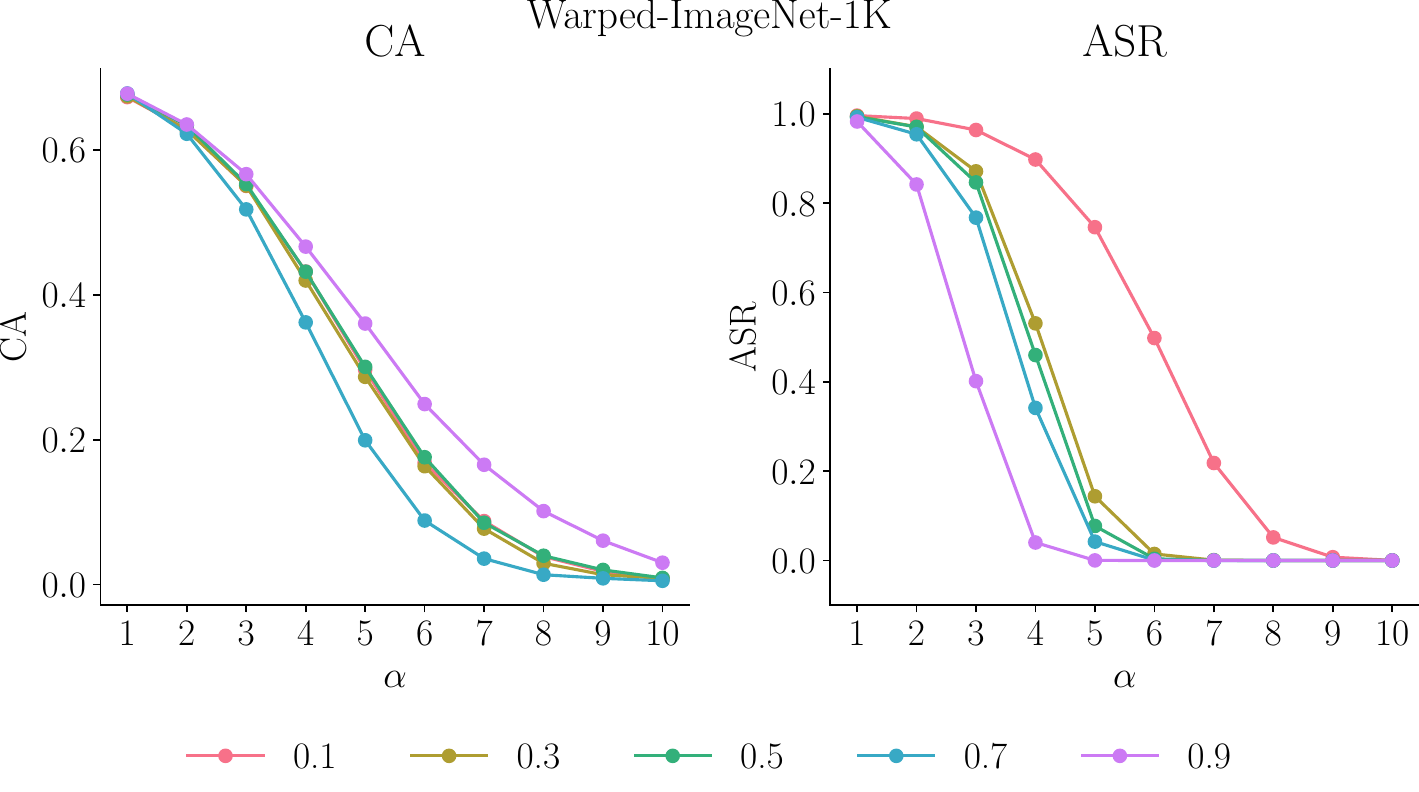}
    \end{subfigure}
    \caption{(ImageNet-1K) Plots showing (CA $\uparrow$) and (ASR $\downarrow$) using task vectors extracted from a mixture of clean and triggered data under varying ratios along increasing scaling values.}
    \label{IN_mixed_ft}
\end{figure}

We also experimented with forget sets with a mixture of clean and triggered data. Figures~\ref{Sun_mixed_ft},~\ref{CIF_mixed_ft},~\ref{IN_mixed_ft}, show the CA and ASR obtained using different ratios of clean:triggered examples in the forget set. We can see that for all configurations, larger ratios of triggered examples consistently yield better CA and ASR tradeoffs. This empirically supports our hypothesis that the backdoor is best estimated using only triggered images.

\subsection{Scaling coefficient sensitivity}

To check if the performance of our method is robust to the choice of scaling coefficient, we present in Table~\ref{tab:stats} sensitivities to this choice within a 10\% variation of the optimal value, averaged over 4 runs of the experiment previously presented in Table~\ref{tab:single-task-negation} of the main text on the WaNet attack. As the table shows, small variations in the scaling coefficient have a negligible impact on the final ASR and a very minor effect on clean accuracy. 

\begin{table}[ht]
\centering
\caption{\looseness=-1 Scaling coefficient sensitivities within a 10\% variation of the optimal value for a single attack run.}
\begin{tabular}{llclc|clc} 
\toprule
\textbf{Dataset} &  & \textbf{-10\% CA} &  & \textbf{\textbf{-10\%~}ASR} & \textbf{\textbf{+10\%~}CA} &  & \textbf{\textbf{\textbf{\textbf{+10\%~}}}ASR}  \\ 
\midrule
SUN397           &  & 73.58 ± 0.27      &  & 0.01 ± 0.01                 & 73.08 ± 0.57               &  & 0.00 ± 0.00                                    \\
CIFAR100         &  & 87.76 ± 0.52      &  & 0.11 ± 0.19                 & 87.39 ± 0.65               &  & 0.03 ± 0.01                                    \\
ImageNet-1K         &  & 66.09 ± 0.94      &  & 0.01 ± 0.01                 & 65.42 ± 1.48               &  & 0.00 ± 0.00                                    \\
\bottomrule
\end{tabular}
\label{tab:stats}
\end{table}

\subsection{More on weight disentanglement}

\begin{figure*}[h]
    \centering
    \includegraphics[width=0.9\linewidth]{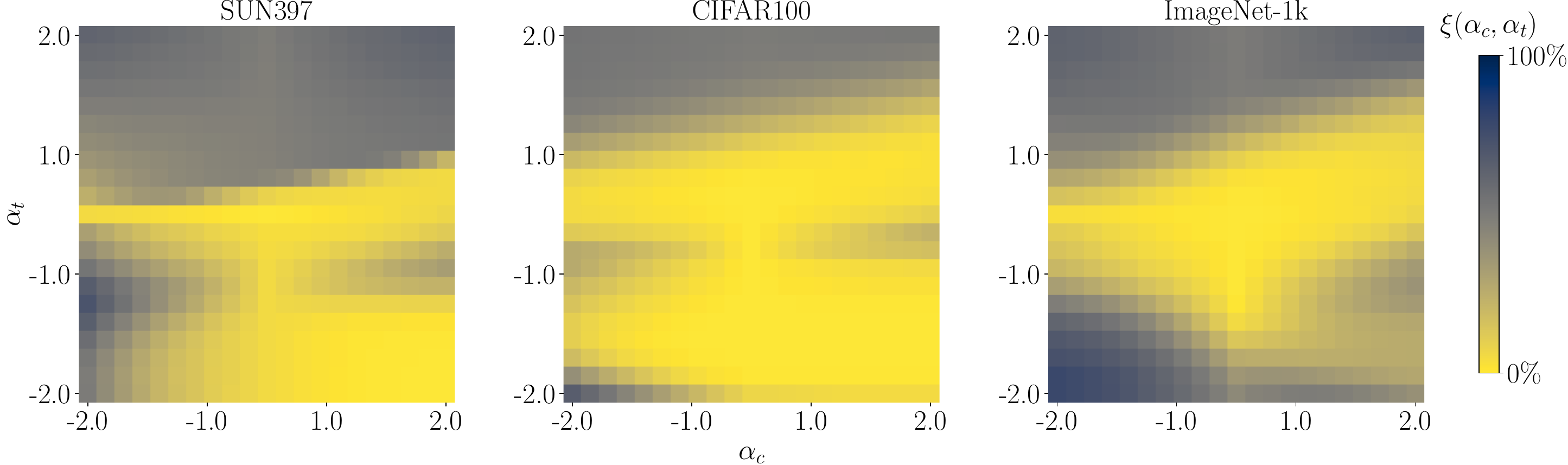}
    \caption{Weight disentanglement between clean and triggered tasks. We estimate the triggered direction $\boldsymbol{\hat \tau}_t$ from the backdoored model and define the clean direction $\boldsymbol{\hat\tau}_c$ as the residual after negation. The plots show the disentanglement error $\xi(\alpha_c, \alpha_t)$ between these task vectors, following \citep{ortiz2024task}. Shown models are backdoored using the \textbf{Blended} attack on the visual encoder of CLIP ViT-B/32.}

    \label{fig:blended disentanglment}
\end{figure*}

\begin{figure*}[h]
    \centering
    \includegraphics[width=0.9\linewidth]{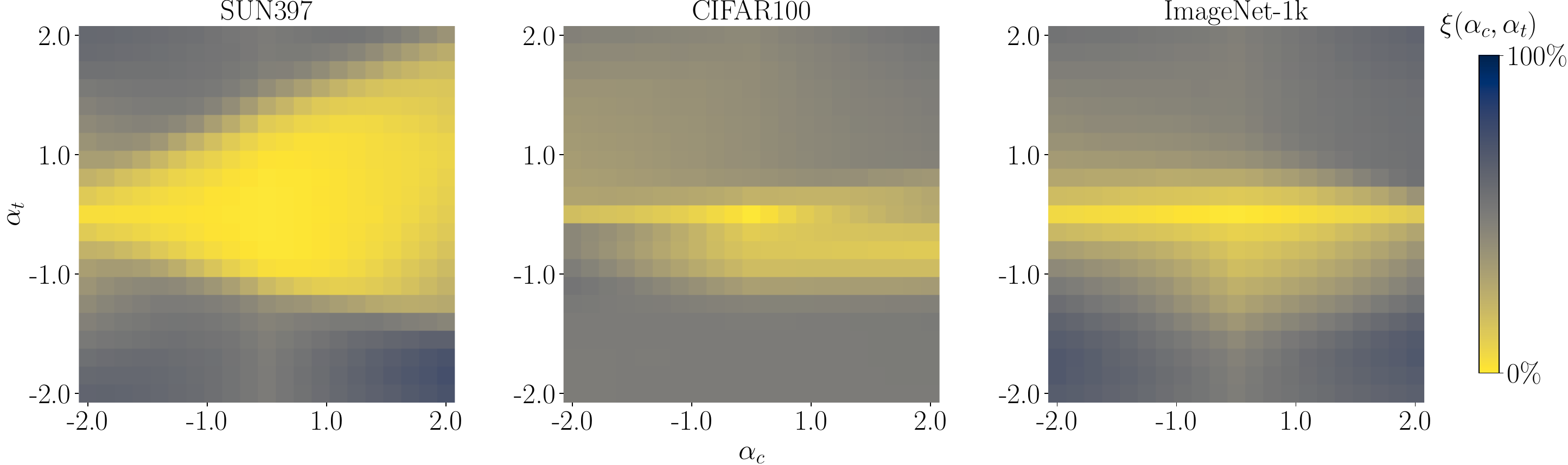}
    \caption{Weight disentanglement between clean and triggered tasks. We estimate the triggered direction $\boldsymbol{\hat \tau}_t$ from the backdoored model and define the clean direction $\boldsymbol{\hat\tau}_c$ as the residual after negation. The plots show the disentanglement error $\xi(\alpha_c, \alpha_t)$ between these task vectors, following \citep{ortiz2024task}. Shown models are backdoored using the \textbf{WaNet} attack on the visual encoder of CLIP ViT-B/32.}

    \label{fig:wanet disentanglment}
\end{figure*}

Figures~\ref{fig:blended disentanglment}, \ref{fig:wanet disentanglment} report additional weight disentanglement visualizations for the attacks considered in Section \ref{sec:analysis}.

\subsection{Additional experiments on other architectures and pre-training}\label{appendix_more_arch}
To further assess the robustness of using \methodName{} across architectures and pre-training settings, we applied our method to:

\begin{itemize}
    \item A convolutional model (ConvNeXt-Base pretrained on LAION‑400M via contrastive learning). See Table~\ref{convnext}.

    \item A transformer model (ViT-B/16) with DINO pre-training on ImageNet-1K backdoored using CIFAR100. See Table~\ref{dino}.
\end{itemize}

\subsection{Detoxifying merged models}\label{appendix_merging}

Recent work by \citet{zhang2024badmerging} examined the behavior of backdoors under model merging, where task vectors from different models are combined directly in parameter space. 

\begin{wraptable}[7]{r}{0.53\textwidth}
\centering
\caption{Unlearning BadMerging \citep{zhang2024badmerging} patches with \methodName{}. Gray denotes ($1-\text{ASR}$).
}
\begin{adjustbox}{width=\linewidth}
\begin{tabular}{@{}llcc|cc@{}}

\toprule
\textbf{} &
& \textbf{CA $\uparrow$} & \textbf{ASR $\downarrow$} 
& \textbf{CA (\methodName{}) $\uparrow$} & \textbf{ASR (\methodName{}) $\downarrow$} \\
\midrule
TA         && 74.02 & 99.66  & 73.50 {\color{gray}(99.30\%)}  & 00.14 {\color{gray}(99.86\%)}  \\
TIES       && 74.96 & 99.92  & 74.54 {\color{gray}(99.44\%)}  & 00.05 {\color{gray}(99.95\%)}  \\
\bottomrule
\end{tabular}
\end{adjustbox}
\label{tab:wrapped-table}
\end{wraptable}
\looseness=-1 They observed that some backdoors fail to persist through merging, leading them to propose BadMerging, a two-stage attack that constructs optimized trigger patches designed to remain functional after merging. Given that BadMerging minimizes its signature in weight space to survive merging, it may similarly resist removal by parameter-space unlearning methods.
Table \ref{tab:wrapped-table} shows the results of applying \methodName{} to models infected with BadMerging and merged using two approaches: Task Arithmetic (TA) \citep{ilharco2022editing}, and TIES \citep{yadav2023ties}, the latter addresses parameter interference through trimming, sign alignment, and selective averaging. \methodName{} substantially reduces the attack success rate in both cases, with minimal degradation in clean accuracy.

\begin{table}[h]
    \centering
    \caption{Controlled experiments showing the effectiveness of \methodName{} on single-task CLIP ConvNeXt-Base classifiers under three backdoor attacks. Clean Accuracy (CA $\uparrow$) and Attack Success Rate (ASR$\downarrow$) are reported before and after unlearning.}
    \label{convnext}
    \begin{tabular}{l|c|c|c|c} 
    \toprule
    \textbf{Dataset}  & \textbf{CA} & \textbf{ASR} & \textbf{CA} (\methodName{}) & \textbf{ASR} (\methodName{})  \\ 
    \midrule
    \rowcolor[HTML]{F7F7F7}\multicolumn{5}{c}{\textit{BadNet}} \\

    CIFAR100 & 89.15    & 99.99     & 82.94    & 02.95      \\
    ImageNet & 72.83    & 99.94     & 67.50    & 02.56      \\
    SUN397   & 76.99    & 99.99     & 67.48    & 05.11      \\

    \rowcolor[HTML]{F7F7F7}\multicolumn{5}{c}{\textit{Blended}}                             \\
   
    CIFAR100 & 89.07    & 99.92     & 87.09    & 00.02      \\
    ImageNet & 72.74    & 99.85     & 71.06    & 00.00      \\
    SUN397   & 76.89    & 99.93     & 73.21    & 00.00      \\

    \rowcolor[HTML]{F7F7F7}\multicolumn{5}{c}{\textit{WaNet}}                               \\
    CIFAR100 & 89.12    & 99.95     & 86.55    & 00.04      \\
    ImageNet & 72.78    & 99.99     & 70.67    & 00.01      \\
    SUN397   & 77.06    & 99.96     & 74.97    & 00.00      \\
    \bottomrule
    \end{tabular}
    \end{table}
    
\clearpage


\begin{table}[h]
    \centering
    \caption{Controlled experiments showing effectiveness of TBAR on transformer model (ViT-B/16) with DINO pre-training on ImageNet-1K under three backdoor attacks using CIFAR100 dataset. Clean Accuracy (CA $\uparrow$) and Attack Success Rate (ASR$\downarrow$) are reported before and after unlearning.}
    \label{dino}
    \begin{tabular}{l|c|c|c|c} 
    \toprule
    \textbf{Attack}  & \textbf{CA} & \textbf{ASR} & \textbf{CA} (\methodName{}) & \textbf{ASR} (\methodName{})  \\ 
    \midrule
    BadNet & 78.98    & 99.63     & 73.98    & 00.11      \\ 
    Blended & 78.74    & 99.34     & 73.30    & 00.00      \\ 
    WaNet & 78.38    & 99.08     & 73.43    & 00.04      \\
    \bottomrule
    \end{tabular}
    \end{table}
    
\vspace{-1.5em}
\section{More Large Scale Experiments} \label{appendix_large_scale}

\subsection{Limitations of clean data finetuning}\label{appendix_ood}

As noted in the main text, large-scale finetuning can cause models to forget broader knowledge. Table~\ref{tab:ood-accuracy} shows performance on SUN397 and CIFAR100 to assess the impact of backdooring and the clean-data baselines from Table~\ref{ViT-Base-IN}. Clean-data finetuning significantly degrades accuracy on these tasks, while \methodName{} has only a minor effect.

\begin{table*}[ht]
\centering
\begin{footnotesize}
\caption{Out-of-distribution clean accuracy on SUN397 and CIFAR100 For CLIP ViT-B/32 model backdoored with image-caption data.
}
\label{tab:ood-accuracy}
\begin{tabular}{@{}lcccccc@{}}
\toprule
\textbf{Dataset} 
& \textbf{Pre-Trained} 
& \textbf{Backdoored} 
& \textbf{CleanCLIP} 
& \textbf{RoCLIP} 
& \textbf{Contrastive-FT} 
& \textbf{\methodName{}} \\
\midrule
\rowcolor[HTML]{F7F7F7}\multicolumn{7}{c}{\textit{BadNet}} \\
SUN397   & 63.18\% & 63.23\% & 56.50\% & 58.47\% & 56.47\% & 61.47\% \\
CIFAR100 & 65.58\% & 63.84\% & 48.38\% & 40.77\% & 52.39\% & 63.89\% \\
\rowcolor[HTML]{F7F7F7}\multicolumn{7}{c}{\textit{Blended}} \\
SUN397   &  63.18\% &  63.19\% &  55.65\% &   56.43\% &  55.60\% &  62.41\% \\
CIFAR100 &  65.58\% &  64.65\% &  52.31\% &  37.91\% & 52.03\% &  64.94\% \\
\rowcolor[HTML]{F7F7F7}\multicolumn{7}{c}{\textit{WaNet}} \\
SUN397   &  63.18\% &  62.84\% & 56.37\% &   55.24\% & 55.66\% &  62.25\% \\
CIFAR100 &  65.58\% &  62.68\% & 53.43\% &  36.32\% &  53.94\% &  61.84\% \\
\bottomrule
\end{tabular}
\end{footnotesize}

\end{table*}

\begin{wraptable}{r}{0.5\textwidth}
\vspace{-1em}
\centering
\caption{Results On CLIP ViT-B/32 with SIG attack, showing (CA $\uparrow$) and (ASR $\downarrow$) performance evaluated on the ImageNet-1K validation set.}
\vspace{0.5em}
{\fontsize{8pt}{9pt}\selectfont
\begin{tabular}{lcc} 
\toprule
           & \multicolumn{2}{c}{SIG}  \\ 
\cmidrule{2-3}
           & CA      & ASR            \\ 
\midrule
Zero-Shot         & 63.34\% & 00.00\%          \\
Backdoored & 61.36\% & 99.01\%        \\
\rowcolor[HTML]{F7F7F7}\multicolumn{3}{l}{} \\
Contrastive-FT         & 51.46\% & 10.26\%        \\
RoCLIP     & 52.61\% & 04.34\%        \\
CleanCLIP  & 51.12\% & 05.51\%        \\ 
\rowcolor[HTML]{F7F7F7}\multicolumn{3}{l}{} \\ 
GA         & 58.25\% & 00.10\%        \\
\methodName{}       & 59.02\% & 00.42\%        \\ 
\rowcolor[HTML]{F7F7F7}\multicolumn{3}{l}{} \\
GA+DECREE      & 56.52\%   &  03.01\%      \\
\methodName{}+DECREE     & 55.41\%   &  05.43\%        \\
\bottomrule
\end{tabular}
}
\end{wraptable}


\vspace{-.5em}

\subsection{Enhancing unlearning robustness with weak trigger cues} 

\looseness=-1 We additionally provide results on unlearning sinusoidal (SIG) attack \citep{barni2019new} on ViT-B/32. In the latter case, we observed that probing the backdoored model with a reverse-engineered SIG patch consistently resulted in the label "television".
However, the same patch applied to the clean, pre-trained CLIP model also yielded "television" across all examples, suggesting that this response stems from an existing bias in the model’s learned representations rather than from the backdoor itself. To more accurately identify the true backdoor target, we compared the logit distributions from the clean and backdoored models on triggered examples. The class with the largest shift in density was indeed the "banana" class. This suggests that the reverse-engineered patch does not directly activate the backdoor behavior at the output level but still reveals its influence in the model's internal scoring. This observation leads to important insights. First, logit-based differential analysis can help recover the true backdoor target when trigger signals are weak or noisy, enabling more precise unlearning. Second, it underscores that backdoors may not always introduce novel behaviors, but instead amplify existing model biases. This logit difference test was additionally evaluated and confirmed for all the experiments reported in the main text.

\end{document}